\documentclass{ieeeaccess}
\usepackage{cite}
\usepackage{amsmath,amssymb,amsfonts}
\usepackage{algorithmic}
\usepackage{graphicx}
\usepackage{textcomp}

\usepackage{subcaption}
\usepackage{rotating}
\usepackage{tabularx}
\usepackage{multirow}
\usepackage{xcolor}
\usepackage{colortbl}
\usepackage{booktabs}

\definecolor{blue1}{HTML}{e6f7ff}
\definecolor{blue2}{HTML}{cceeff}
\definecolor{blue3}{HTML}{b3e6ff}

\definecolor{red1}{HTML}{fff0e6}
\definecolor{red2}{HTML}{ffe0cc}
\definecolor{red3}{HTML}{ffd1b3}
\definecolor{red4}{HTML}{ffc299}
\definecolor{red5}{HTML}{ffb380}
\definecolor{red6}{HTML}{ffa366}

\definecolor{grey}{HTML}{f2f2f2}

\begin{document}
\history{Date of publication xxxx 00, 0000, date of current version xxxx 00, 0000.}
\doi{10.1109/ACCESS.2017.DOI}

\title{Visual Probing: Cognitive Framework for Explaining Self-Supervised Image Representations}
\author{\uppercase{Witold~Oleszkiewicz}\authorrefmark{1},
\uppercase{Dominika~Basaj}\authorrefmark{1,5}, \uppercase{Igor~Sieradzki}\authorrefmark{2}, \uppercase{Michal~Gorszczak}\authorrefmark{2}, \uppercase{Barbara~Rychalska}\authorrefmark{1,4}, \uppercase{Koryna~Lewandowska}\authorrefmark{3}, \uppercase{Tomasz~Trzcinski}\authorrefmark{1,2,5}, \IEEEmembership{Member, IEEE}, \uppercase{and Bartosz~Zielinski}.\authorrefmark{2,6}.}
\address[1]{Warsaw University of Technology, plac Politechniki 1, Warszawa, Poland}
\address[2]{Faculty of Mathematics and Computer Science, Jagiellonian University, Łojasiewicza 6, Kraków, Poland}
\address[3]{Cognitive Neuroscience and Neuroergonomics, Institute of Applied Psychology, Łojasiewicza 4, Kraków, Poland}
\address[4]{Synerise, Lubostroń 1, Kraków, Poland}
\address[5]{Tooploox, Tęczowa 7, Wrocław, Poland}
\address[6]{Ardigen, Podole 76, Kraków, Poland}
\tfootnote{This research was funded by Foundation for Polish Science (grant no POIR.04.04.00-00-14DE/18-00 carried out within the Team-Net program co-financed by the European Union under the European Regional Development Fund), National Science Centre, Poland (grant no 2020/39/B/ST6/01511). The authors have applied a CC BY license to any Author Accepted Manuscript (AAM) version arising from this submission, in accordance with the grants’ open access conditions. Dominika Basaj was financially supported by grant no 2018/31/N/ST6/02273 funded by National Science Centre, Poland.}

\markboth
{Witold Oleszkiewicz \headeretal: Visual Probing: Cognitive Framework for Explaining Self-Supervised Image Representations}
{Witold Oleszkiewicz \headeretal: Visual Probing: Cognitive Framework for Explaining Self-Supervised Image Representations}

\corresp{Corresponding author: Witold Oleszkiewicz (e-mail: witold.oleszkiewicz@pw.edu.pl).}
.

\begin{abstract}
Recently introduced self-supervised methods for image representation learning provide on par or superior results to their fully supervised competitors, yet the corresponding efforts to explain the self-supervised approaches lag behind. Motivated by this observation, we introduce a novel visual probing framework for explaining the self-supervised models by leveraging probing tasks employed previously in natural language processing. The probing tasks require knowledge about semantic relationships between image parts. Hence, we propose a systematic approach to obtain analogs of natural language in vision, such as visual words, context, and taxonomy. Our proposal is grounded in Marr's computational theory of vision and concerns features like textures, shapes, and lines. We show the effectiveness and applicability of those analogs in the context of explaining self-supervised representations. Our key findings emphasize that relations between language and vision can serve as an effective yet intuitive tool for discovering how machine learning models work, independently of data modality. Our work opens a plethora of research pathways towards more explainable and transparent AI.
\end{abstract}


\titlepgskip=-15pt

\maketitle

\section{Introduction}

\begin{figure*}
\centering
\includegraphics[width=0.8\linewidth]{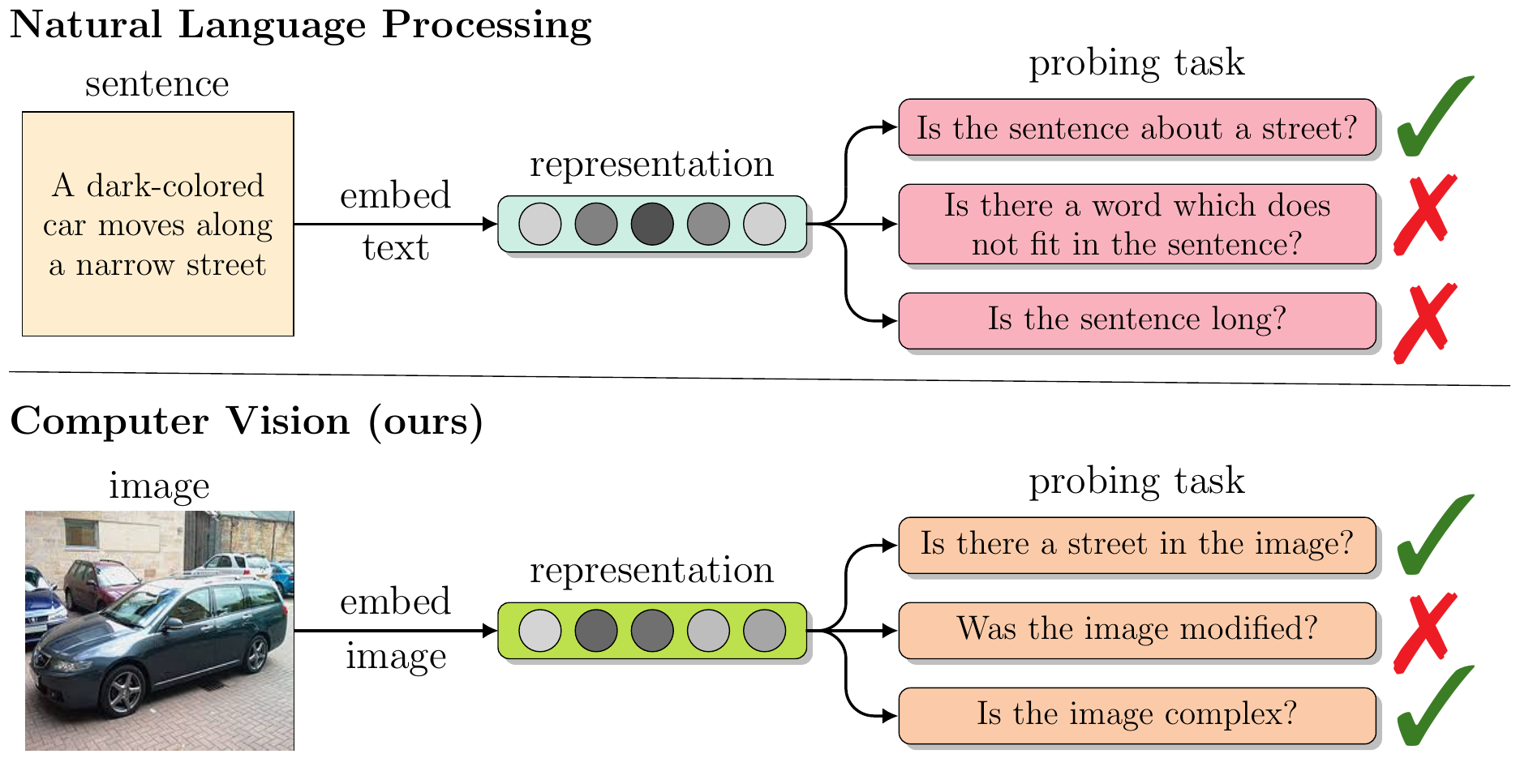}
\caption{{\it Probing tasks}, widely used in natural language processing, validate if a {\it representation} implicitly encodes a given property, {\it e.g.} a sentence topic or its length. We introduce a visual taxonomy along with the corresponding probing framework that allows to build analogous {\it visual probing tasks} and explain the self-supervised image representations. As a result, we \textit{e.g.} discover that the information stored by self-supervised representations is biased towards lines and forms than textures.}
\label{fig:idea}
\end{figure*}

\PARstart{V}{isual} representations are cornerstones of a multitude of contemporary computer vision and machine learning applications, ranging from visual search~\cite{sivic2006video} to image classification~\cite{krizhevsky2012imagenet} and Visual Question Answering, VQA~\cite{antol2015vqa}. However, learning representations from data typically requires tedious annotation. Therefore, recently introduced self-supervised representation learning methods concentrate on decreasing the need for data labeling without reducing their performance~\cite{chen2020big,grill2020bootstrap,caron2020unsupervised}. Because of the fundamental role representations play in real-life applications, much research focuses on explaining these embeddings~\cite{vulic2020probing,eichler2019linspector,huang2020interpretable}. Nevertheless, most of them concentrate on fully supervised embeddings~\cite{zhang2018visual} and not on their self-supervised counterparts. Moreover, the majority of the proposed approaches rely on pixel-wise image analysis~\cite{simonyan2013deep,adebayo2018sanity}, while general semantic concepts present in the images are often ignored.

Here, we attempt to overcome these shortcomings and draw inspiration from a simple yet often overlooked observation that humans use language as a natural tool to explain what they learn about the world through their eyes~\cite{kumar2020nile}. Therefore, considering that the very same machine learning algorithms can be successfully applied to solve both vision and Natural Language Processing (NLP) tasks~\cite{dosovitskiy2020image,carion2020end}, we postulate that the methods used to analyze text representation can also be employed to investigate visual inputs.

Very popular tools for explaining textual embeddings are {\it probing tasks}~\cite{conneau2018you}. As shown in the upper part of Figure~\ref{fig:idea}, a probing task in NLP is a simple classifier that asks if a given textual representation encodes a particular property, such as a sentence length or its semantic consistency, even though this property was not a direct training objective. 
For instance, we can create a textual probing task by substituting a word in a sentence and checking if a simple classifier that takes the representation of the original and altered sentence can detect this change. Furthermore, by analyzing the accuracy of a probing task, one can verify if the investigated representation contains certain information and understand the rationale behind embedding creation. However, while probing tasks are straightforward, intuitive, and widely used tools in NLP, their computer vision application is limited~\cite{alain2016understanding}, mainly due to the lack of appropriate analogs between textual and visual modalities.

In this paper, we address this limitation by introducing an intuitive mapping between vision and language that enables applying the NLP probing tools in the computer vision (CV) domain. For this purpose, in Section~\ref{sec:visual_probing}, we propose a taxonomy of visual units that includes \textit{visual sentences}, \textit{words}, and \textit{characters}. We describe them using visual features presented in Marr's computational theory of vision~\cite{Marr:1982:VCI:1095712}, such as texture, shapes, and lines. Finally, we employ them as building blocks for a more general visual probing framework that contains a variety of NLP-inspired probing tasks, such as \textit{Word Content}, \textit{Sentence Length}, \textit{Character Bin}, and \textit{Semantic Odd Man Out}~\cite{conneau2018you,eichler2019linspector}.
The results we obtain provide us with unprecedented insights into semantic knowledge, complexity, and consistency of self-supervised image representations, {\it e.g.} we discover that semantics of the image only partially contribute to target task accuracy. One of our key findings is that the information stored by self-supervised representations is much more influenced by lines and forms than textures. What confirms the design choices behind hand-crafted visual representations such as SIFT~\cite{Lowe04} or BRIEF~\cite{Calonder12}. Our framework also allows us to compare the existing self-supervised representations from a novel perspective, as we show in Section~\ref{sec:results}.

Our contributions can be therefore summarized as follows:
\begin{itemize}
    \item We propose an intuitive mapping between visual and textual modalities that constructs a visual taxonomy.

    \item We introduce novel visual probing tasks for comparing self-supervised image representations inspired by similar methods used in NLP.

    \item We show that leveraging the relationship between language and vision serves as an effective yet intuitive tool for discovering how self-supervised models work.
\end{itemize}

\section{Related Works}
\label{sec:related_works}

The visual probing framework aims to explain image representations obtained from self-supervised methods. Moreover, it is inspired by probing tasks used in NLP. Therefore, in this section, we consider related works from three research areas: self-supervised computer vision models, probing tasks in natural language processing, and explainability methods in computer vision.

\subsection{Self-supervised computer vision models} 
Earliest self-supervised methods were based on a  pretext task, for example image colorization \cite{zhang2016colorful} or rotation prediction \cite{DBLP:journals/corr/abs-1803-07728} using cross-entropy loss. However, recently published state-of-the-art methods usually base on contrastive loss~\cite{hadsell2006dimensionality}, which measures the similarities of patches in representation space and aims to discriminate between positive and negative pairs. The positive pair contains modified versions of the same image, while the negative pairs correspond to two images in the same dataset.
One of the methods, called MoCo~v1~\cite{he2020momentum} trains a slowly progressing encoder, driven by a momentum update. This encoder plays the role of a large memory bank of past representations and delivers information about negative examples.
Another method, called SimCLR~v2~\cite{chen2020big}, proposes a different way of generating negative pairs, using a large batch size of up to 4096 examples. Other important improvements proposed by SimCLR~v2 are the projection head and carefully tuned data augmentation. The projection head maps representations into space where contrastive loss is applied to prevent the loss of information.
On the other hand, BYOL~\cite{grill2020bootstrap} also uses the projection head, but unlike MoCo~v1 and SimCLR~v2, it achieves a state-of-the-art performance without the explicitly defined contrastive loss function, so it does not need negative examples.
Finally, SwAV~\cite{caron2020unsupervised} first obtains ``codes'' by assigning features to prototype vectors, and then solves a ``swapped'' prediction problem wherein the codes obtained from one data augmented view are predicted using the other view.
Our paper provides a framework for analyzing the representations generated by those methods regarding the semantic knowledge they encode.

\subsection{Probing tasks in NLP} NLP probing tasks aim to probe word or sentence representations for interesting linguistic features to discover whether they contain linguistic knowledge~\cite{DBLP:journals/tacl/BelinkovG19}. Probing is usually achieved with a binary or multi-class classifier, which takes one or two-word embeddings as input, and predicts the existence or absence of a chosen linguistic phenomenon in the input representation(s) \cite{conneau2018you}. The qualities of a good probing classifier are the subject of a debate, as too expressive probes could learn important features on their own, even if the information is not present in the representations \cite{hewitt2019designing}. Thus, probing is usually achieved with simple classifiers. 

Classic probing literature considers various linguistic aspects, from the simplest to very complex ones. In ~\cite{conneau2018you} the probed linguistic features are, for example, the depth of the sentence parse tree or whether the sentence contains a specific word. Other works propose to focus on lexical knowledge concerning the qualities of individual words more than the whole sentences~\cite{vulic2020probing,eichler2019linspector}, probing token embeddings for qualities such as gender, case, and tense, or differentiation between real words and pseudowords~\cite{eichler2019linspector}. Other approaches focus on certain kinds of words, e.g., function words, such as \textit{wh}-words and propositions   \cite{kim-etal-2019-probing}. We consider all these objectives in our approach, i.e., we study probing tasks on both individual concepts and their compositions. Moreover, while most works on probing tasks focus on one selected language, the others~\cite{eichler2019linspector} are designed with multilingual settings in mind. It has been shown that it is possible to create NLP probing tasks that are transferable across languages, even if the languages vary considerably in their structure, which means that probing tasks can touch upon more universal cognitive phenomena  \cite{krasnowska-kieras-wroblewska-2019-empirical}. This paper also aims at the flexibility and universality of our probing tasks, as our approach can be applied to various image domains.

\subsection{Explainability methods in CV representation learning}

eXplainable Artificial Intelligence (XAI) gains popularity fuelled by the black-box character of today's deep neural networks~\cite{Ribeiro2016, ghorbani2019towards}.
Popular explainability approaches for model explanations are saliency or attention maps, which provide the importance of weights to pixels based on the first order derivatives~\cite{Mahendran2014,simonyan2013deep,adebayo2018sanity,chattopadhyay2017} but do not fully explain the reasoning behind the actual decision~\cite{sixt2019explanations} and do not describe the concrete semantic concepts. Moreover, some of the methods are even agnostic of the model itself~\cite{adebayo2018sanity} and thus are not able to explain it. Another common local approach is perturbation-based interpretability, which applies changes to either data~\cite{landesberger_visual_2019} or features~\cite{ribeiro2016should} and observes the influence on the output.

Some methods verify the relevance of network hidden layers. For example, \cite{alain2016understanding} use linear classifiers trained on representations from these layers to measure how suitable they are for the classification. Subsequent efforts focused on understanding the function of hidden layers led to the introduction of network dissection~\cite{Bau2017, Zhou2017}, which enables quantifying the interpretability of latent representations by evaluating the alignment between their hidden units and a set of visual semantic concepts obtained from human annotators.

More recent methods are inspired by the human brain and how it explains its visual judgments by pointing to prototypical features that an object possesses~\cite{salakhutdinov2012one}. I.e., a certain object is a car because it has tires, a roof, headlights, and a horn. For example, prototypical part network~\cite{Chen2018} applies this paradigm by focusing on parts of an image and comparing them with prototypical parts of a given class. At the same time, the extension proposed in~\cite{rymarczyk2021protopshare} uses data-dependent merge-pruning of the prototypes to allow sharing them among the classes.
Another promising approach is Concept Activation Vector (CAV), defined in the feature space to quantify the degree to which a predefined concept is vital for a prediction~\cite{kim2018interpretability}. This approach has recently been extended to automatically discovered concepts~\cite{ghorbani2019towards} and to interactive techniques used by pathologists to indicate what characteristics are essential when searching for similar images~\cite{cai2019human}.
We propose to continue and extend this line of research by introducing visual word probing, which systematically explains the self-supervised representations.

\section{Visual Probing}
\label{sec:visual_probing}

\begin{figure*}[t!]
\centering
\includegraphics[width=0.95\linewidth]{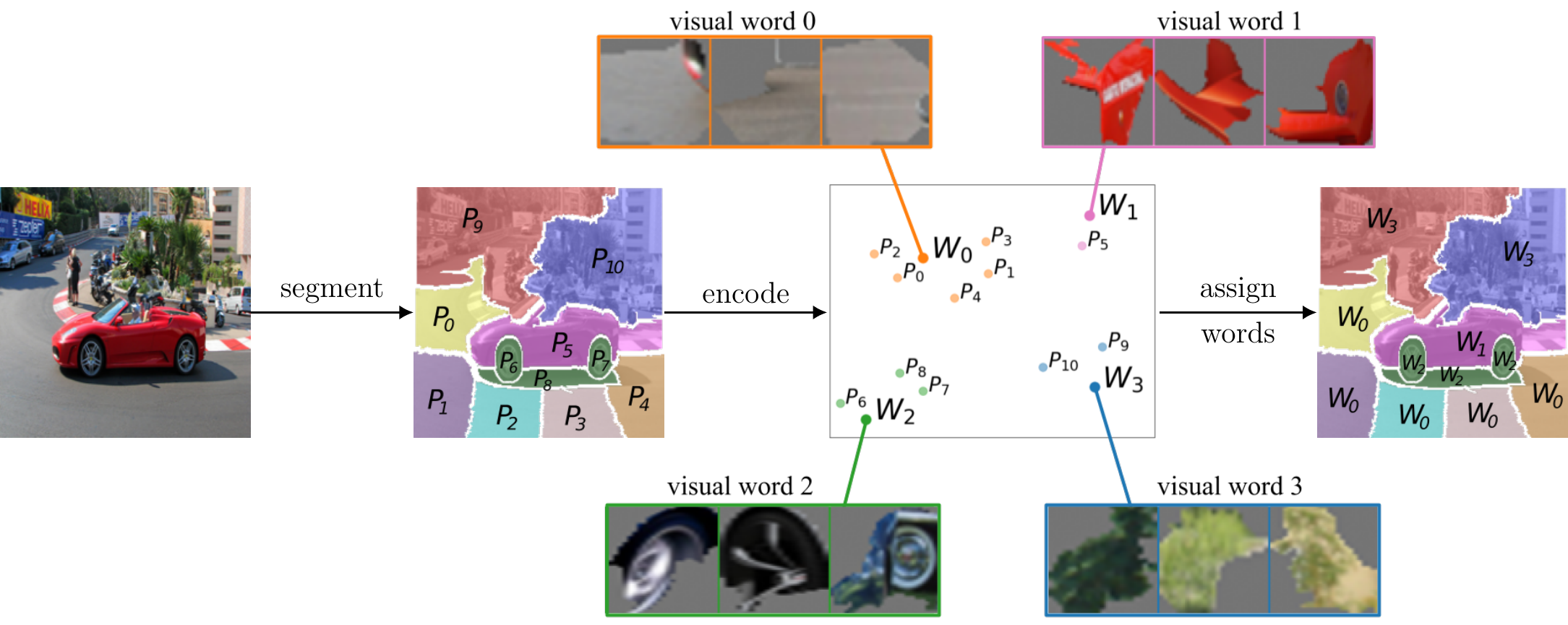}
\caption{The process of dividing an image into visual words. First, an image is segmented into multiple superpixels: $P_0$, $P_1$, $\ldots$, $P_{10}$. Then, each superpixel is embedded in the latent space previously used to generate the dictionary of visual words: $W_0$, $W_1$, $W_2$, $W_3$. Finally, each superpixel is assigned to the closest word in the visual word dictionary. This results in mapping between vision and language and enables using the visual probing framework that includes a variety of NLP-inspired probing tasks.}
\label{fig:visual_words}
\end{figure*}

This section introduces a novel visual probing framework that analyzes the information stored in self-supervised image representations. For this purpose, in Section~\ref{subsec:nlp_in_cv}, we propose a mapping between visual and textual modalities that constructs a visual taxonomy. As a result, the image becomes a ``visual sentence'' constructed from ``visual words'' and can be analyzed with visual probing tasks inspired by similar methods used in NLP (see Section~\ref{subsec:probing_tasks}). Moreover, for in-depth analysis of the concepts trained by self-supervised methods, in Section~\ref{subsec:marr}, we provide a cognitive visual systematic that identifies a visual word with structural features from Marr's computational theory~\cite{Marr:1982:VCI:1095712}.

\subsection{Mapping Between Vision and NLP}
\label{subsec:nlp_in_cv}

After defining the images as analogous to sentences within our framework, the question remains which parts of an image should be considered as equivalent to individual words and characters?
There are multiple possible answers to this question. One of the intuitive ones is to divide an image into non-overlapping superpixels that group pixels into perceptually meaningful atomic regions~\cite{achanta2012slic}. As a result, we obtain an image built from superpixels, an analogy of a sentence built from the words. The superpixels, similarly to words, have their order and meaning (see Section~\ref{subsec:marr}). Moreover, each superpixel contains a specific number of pixels, like the number of characters in a word. As a consequence, we obtain an intuitive mapping between visual and textual domains.

However, superpixels differ conceptually from their linguistic counterparts in one important aspect: they do not repeat between different images, while in text, the words often repeat between sentences.
Therefore, we propose to define visual words as the clusters of all training superpixels in representation space and assign each superpixel to the closest centroid from such a dictionary. For this purpose, we could use the original definition of visual words from~\cite{leung2001representing}. However, it does not take into consideration the importance of those words for a model's prediction. Therefore, we use TCAV methodology~\cite{kim2018interpretability,ghorbani2019towards} that generates high-level concepts, which are important for prediction and easily understandable by humans. Such an approach requires a supervisory training network but generates visual words independent of any of the analyzed self-supervised techniques, which is crucial for a fair comparison. To summarize, the process of dividing an image into visual words consists of three steps: segmentation into superpixels, their encoding, and assignment to visual words (see Figure~\ref{fig:visual_words}).

\subsection{Cognitive Visual Systematic}
\label{subsec:marr}

In contrast to words in NLP, visual words do not have a well-defined meaning required for in-depth analysis of self-supervised representations. Hence, in this section, we introduce cognitive visual systematic, considering that generating visual words is similar to the process of concept formation. This process, described in psychology and cognitive science, is traditionally understood as an internal cognitive representation of a set of similar objects, i.e., ``an idea that includes all that is characteristically associated with it''~\cite{Medin1989-MEDCAC}. In other words, concepts are created in relation to features that constitute similarity amongst included objects.

What features could then be the basis for the formation of visual words? Reference to Marr's computational theory of vision \cite{Marr:1982:VCI:1095712,10.2307/187817} seems to be an appropriate aid in an attempt to answer this question. Marr assumed that perception is achieved by detecting an object's specific structural features, which are then organized in a series of visual representations. Among those, three constitute the major representations: the ``primal sketch'', the ``2.5D sketch'' and the 3D model representation'' \cite{Marr:1982:VCI:1095712}. The primal sketch is a two-dimensional image that uses information on light intensity changes, featuring blobs, edges, lines, boundaries, bars, and terminations. Colors and textures are also thought to be detected on this level \cite{GUO20075,MORGAN2011738}. The 2.5D sketch represents mostly two-dimensional shapes and their orientation towards a viewer-centered location (the sense of image depth is achieved in this stage \cite{10.2307/187817}). Finally, the 3D model is a representation suitable for object recognition. In this stage, the observer can imagine the object from different views. This includes surfaces that are currently invisible to the observer~\cite{Marr:1982:VCI:1095712,10.2307/187817}.
    
\begin{figure}[t!]
    \begin{subfigure}{0.9\linewidth}
    \centering
    \includegraphics[width=\linewidth]{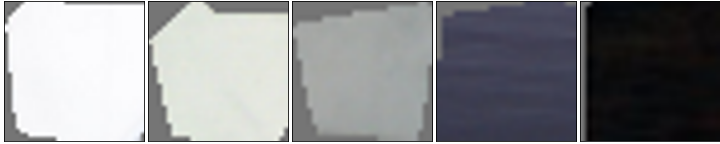}
    \caption{Brightness.}
    \end{subfigure}
    \hfill
    \begin{subfigure}{0.9\linewidth}
    \centering
    \includegraphics[width=\linewidth]{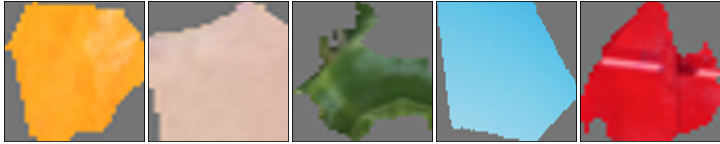}
    \caption{Color.}
    \end{subfigure}
    \hfill
    \begin{subfigure}{0.9\linewidth}
    \centering
    \includegraphics[width=\linewidth]{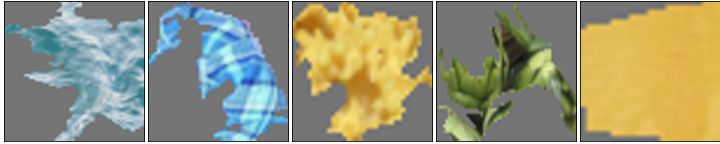}
    \caption{Texture.}
    \end{subfigure}
    \hfill
    \begin{subfigure}{0.9\linewidth}
    \centering
    \includegraphics[width=\linewidth]{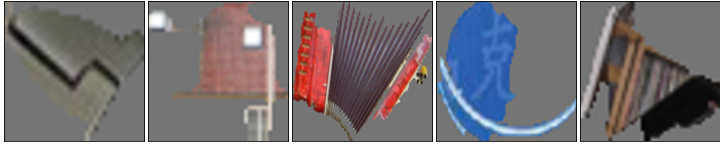}
    \caption{Lines.}
    \end{subfigure}
    \hfill
    \begin{subfigure}{0.9\linewidth}
    \centering
    \includegraphics[width=\linewidth]{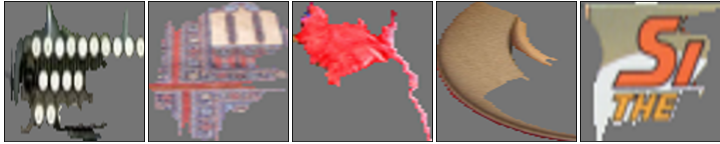}
    \caption{Shape.}
    \end{subfigure}
    \hfill
    \begin{subfigure}{0.9\linewidth}
    \centering
    \includegraphics[width=\linewidth]{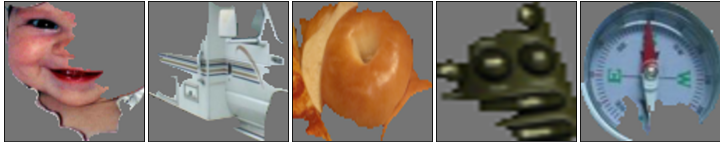}
    \caption{Form.}
    \end{subfigure}
    \caption{Sample superpixels illustrating six visual concepts from the Marr's computational theory of vision.}
    \label{fig:marr}
\end{figure}

To simplify visual word description in terms of Marr’s theory, we decided to use concepts of light intensity (brightness), color, texture, and lines in relation to primal sketch, shape in relation to 2.5D sketch, and form in relation to 3D model (examples are depicted in Figure~\ref{fig:marr}). We use enlisted concepts in the user study to establish the meaning of the visual words we use.

\subsection{Visual Probing Tasks}
\label{subsec:probing_tasks}

After dividing an image into visual words, its representation can be analyzed by the visual probing framework that can adapt most NLP probing tasks. Here, we describe adaptations of five of them, including those well known in the NLP community~\cite{conneau2018you,eichler2019linspector} together with their original NLP definitions to make the paper self-contained.

\subsubsection{Word Content (WC)}
\begin{figure}[t!]
    \centering
    \includegraphics[width=0.9\linewidth, keepaspectratio]{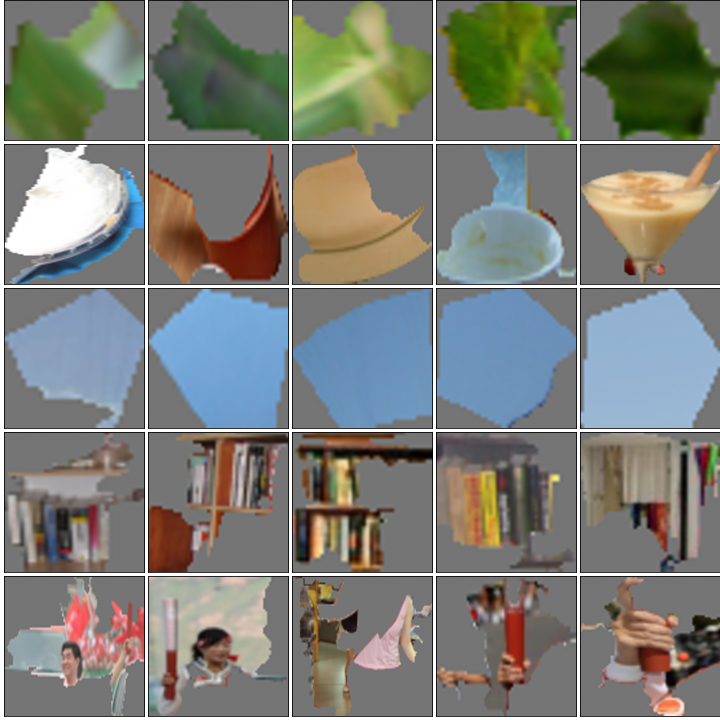}
    \caption{Sample visual words, each represented by one row of five superpixels.}
    \label{fig:wc}
\end{figure}

The \textit{Word Content} probing task aims to identify which visual words are present in an image (see Figure~\ref{fig:wc}). The \textit{input} of this probing task is a self-supervised representation of the image. The \textit{target labels} represent the presence of a particular visual word. As we describe in Section~\ref{sec:experimental_setup}, all visual words are clustered into 50 clusters. Hence, there are 50 binary \textit{target labels}.
Figure~\ref{fig:visual_words} illustrates the process of determining which visual words are present in the image. The NLP inspiration of the task probes for surface information, i.e. the type of information that does not require any linguistic knowledge~\cite{conneau2018you}. In contrast, its adaptation requires \textit{semantic knowledge} to understand which concept is represented by a superpixel.

\subsubsection{Sentence Length (SL)}
\begin{figure}[t!]
\centering
    \includegraphics[width=0.9\linewidth, keepaspectratio]{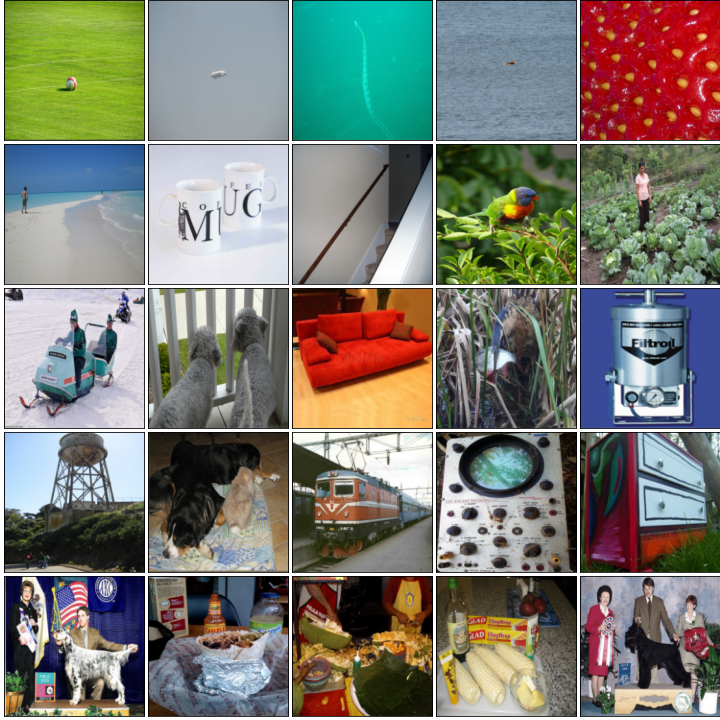}
    \caption{Sample images grouped into rows with increasing value of \textit{Sentence Length} from top to bottom. One can observe that SL correlates with the semantic complexity of the image.}
\label{fig:sentence_length}
\end{figure}

The aim of the \textit{Sentence Length} probing task is to distinguish between simple and complex images, as presented in~Figure~\ref{fig:sentence_length}. The \textit{input} of this probing task is a self-supervised representation of the image. The \textit{target label} is the number of unique visual words in the image, which can be determined based on the WC labels. The original NLP probing task predicts the number of words (or tokens) and retains only surface information~\cite{conneau2018you}. In CV, it serves as a proxy for \textit{semantic complexity}, requiring the semantic understanding of the image.

\subsubsection{Character Bin (CB)}
\begin{figure}[t!]
    \begin{subfigure}{0.9\linewidth}
    \centering
    \includegraphics[width=\linewidth]{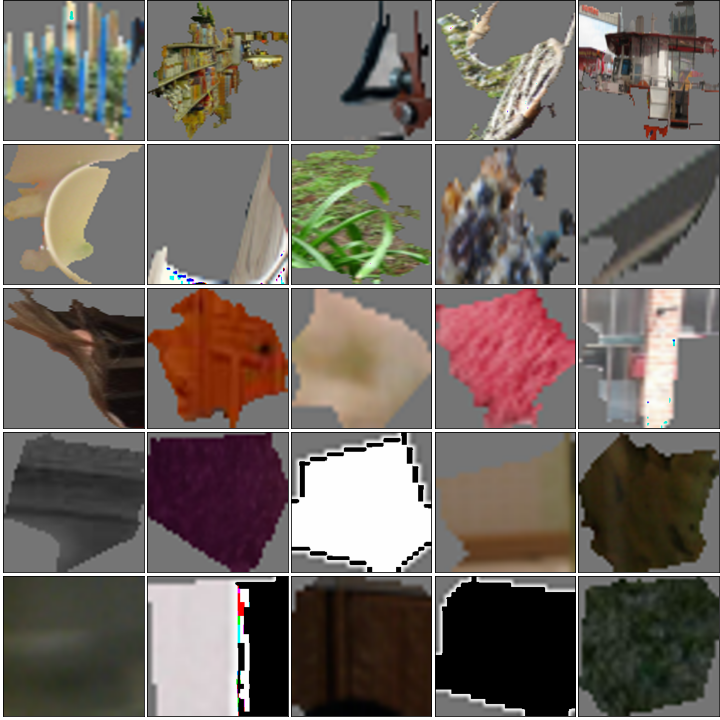}
    \caption{\textit{Character Bin} shape. }
    \label{fig:cb_shape}
    \end{subfigure}\hfill%
    \begin{subfigure}{0.9\linewidth}
    \includegraphics[width=\linewidth]{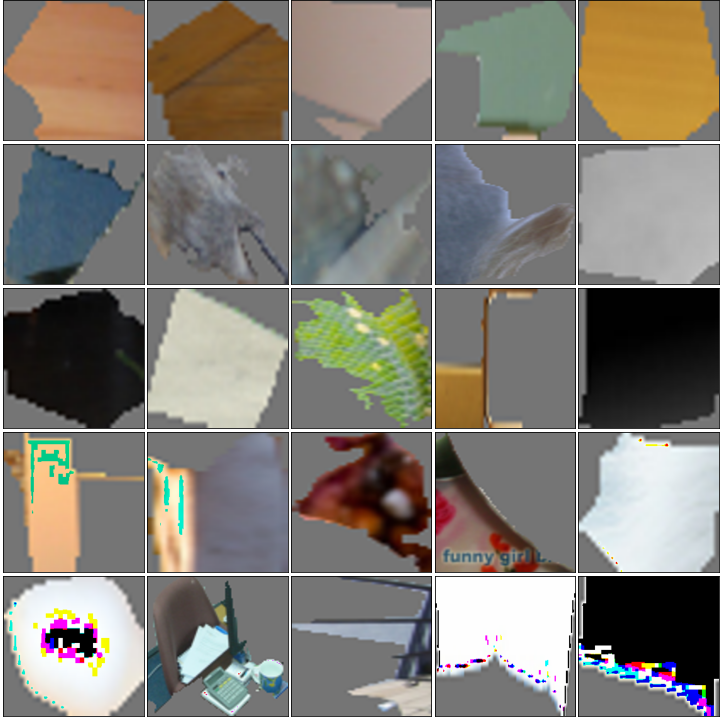}
    \caption{\textit{Character Bin} color. }
    \label{fig:cb_color}
    \end{subfigure}
    \caption{Sample superpixels grouped into rows with increasing values of $CO$ (a) and $ICV$ (b) from top to bottom. One can observe that bottom rows contain superpixels of rounder shape (a) and higher contrast (b).}
    \label{fig:superpixels}
\end{figure}

The aim of the \textit{Character Bin} probing task is to check whether the representation stores information about the complexity of the visual word represented by a superpixel. The \textit{input} of this probing task is a self-supervised representation of the image’s superpixel, and we define two \textit{target labels} that are commonly used in CV literature to describe superpixels. The first target label is the compactness (CO)~\cite{compactness} of the superpixel $S$ defined as the area of the superpixel $A(S)$ divided by the area $A(C)$ of a circle $C$ with the same perimeter as $S$:
$$
CO(S)=\frac{A(S)}{A(C)}.
$$
Sample superpixels with various ranges of CO are presented in~Figure~\ref{fig:cb_shape}.
The second target label is Intra-Cluster Variation (ICV) ~\cite{icv}  defined as the average standard deviation $\sigma_c(S)$ of channels $c\in C$ for superpixel $S$:
$$
ICV(S) = \frac{1}{\left | C \right |} \sum_{c\epsilon C} \sigma_c(S).
$$
Sample superpixels with various ranges of ICV are in~Figure~\ref{fig:cb_color}.
The original NLP probing task is defined as a classifier of the number of characters in a single word~\cite{eichler2019linspector}. From this perspective, the \textit{Character Bin} retains only surface information in both domains.

\subsubsection{Semantic Odd Man Out (SOMO)}
\begin{figure}[t]
    \centering
    \begin{subfigure}{0.9\linewidth}
    \centering
    \includegraphics[width=\linewidth]{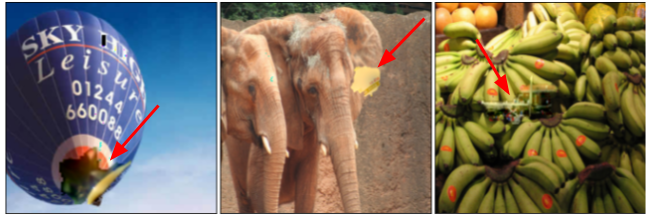}
    \caption{SOMO far.}
    \end{subfigure}\hfill%
    \begin{subfigure}{0.9\linewidth}
    \centering
    \includegraphics[width=\linewidth]{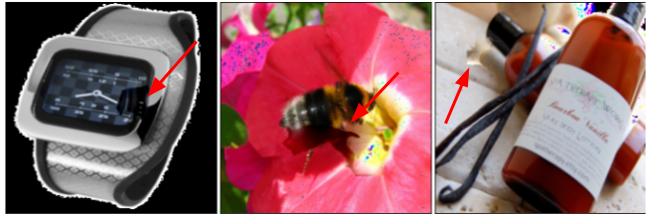}
    \caption{SOMO close.}
    \end{subfigure}
    \caption{Sample images from SOMO far (a) and close (b) setup. Replacements in SOMO close come from a set of visual words that often co-occur with the replaced visual word. In SOMO far, we replace superpixel with rarely co-occurring visual words. One can observe that in the case of SOMO close, the differences are less visible. Notice that red arrows indicate alterations to the image.}
    \label{fig:somo}
\end{figure}

The objective of the SOMO probing task is to predict whether the image was modified. We replace a center-biased superpixel in the image with a similarly shaped superpixel from another image that corresponds to different visual words. We pick a superpixel by using a two-dimensional Gaussian distribution located in the middle of the image. When it comes to replacement, we consider two setups, SOMO close and far, depending on how often two visual words co-occurred in the training images. In SOMO close, we replace a center-biased superpixel with visual words that often co-occur with the replaced visual word. In SOMO far, we replace superpixel with the rarely co-occurring visual word (see Figure~\ref{fig:somo}). In both cases, the \textit{input} of the probing task is a self-supervised representation of the image. The \textit{target label} is binary, i.e., the image was modified or not. The original NLP task predicts if replacing a random noun or verb alters the sentence~\cite{conneau2018you}. In both domains, it requires the ability to detect alterations in \textit{semantic consistency}.

\subsubsection{Mutual Word Content (MWC)}

\begin{figure}[t!]
    \centering
    \begin{subfigure}{0.9\linewidth}
         \centering
         \includegraphics[width=\textwidth]{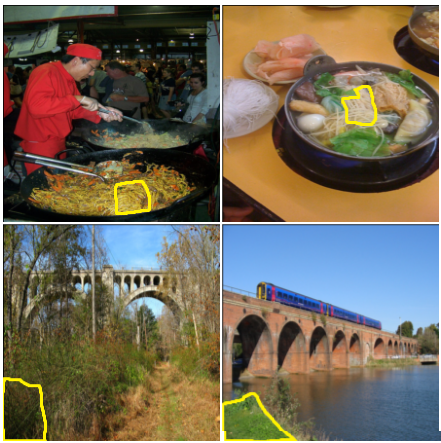}
         \caption{Superpixels that attracts representations.}
     \end{subfigure}
     \hfill
     \begin{subfigure}{0.9\linewidth}
         \centering
         \includegraphics[width=\textwidth]{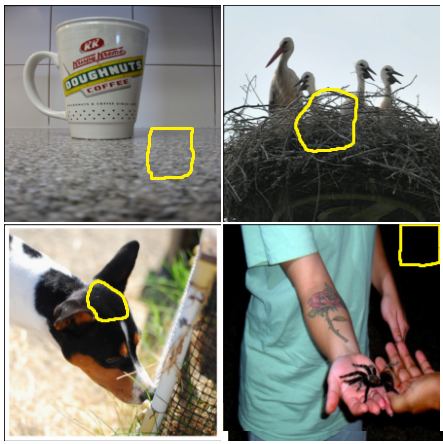}
         \caption{Superpixels that push representations away.}
     \end{subfigure}
    \caption{Sample pairs of images with visual words (represented by marked superpixels) that attract (a) or push away (b) the representations. Visual words that attract representations include words with complicated forms and ``green'' words. On the other hand, visual words that push representations away contain fine textures.}
    \label{fig:mwc}
\end{figure}

The \textit{Mutual Word Content} (MWC) probing task aims to discover which visual words bring two self-supervised representations close to each other and which ones push them farther away (see Figure~\ref{fig:mwc}). The \textit{input} of this probing task is a pair of self-supervised representations of two images. The \textit{target labels} represent the presence of a particular visual word in both images.
The probing task classifier is validated on equally-sized subsets $\{S_{val}^{i}\}$ corresponding to the increasing cosine distance between pairs of representations. Discrepancies in the classifier accuracy show the impact of visual words on the representations’ distance. More precisely, if the probing task performance drops with increasing representations’ distance, the visual word information in both representations brings them closer. To quantify this relationship, we introduce the attraction coefficient. To calculate this coefficient, we use the Linear Regression fit to the points $(i, AUC_i)$, where $i$ is the index of subset and $AUC_i$ is the MWC probing task performance on this subset. Thus, the attraction coefficient is the first derivative of the fitted model.
This probing task does not have a direct counterpart in the NLP domain.

\section{Experimental Setup}
\label{sec:experimental_setup}

This section describes how we generate visual words and self-supervised representations, assign visual words to images, and train the probing tasks\footnote{The code available at: \underline{github.com/BioNN-InfoTech/visual-probes}}.

\subsection{Generating visual words}
\label{subsec:generate_visual_words}
This paper presents a general framework that can be used with various methods of generating visual words. However, choosing a high-quality method is crucial to draw meaningful conclusions from the probing tasks. That is why we use the established ACE algorithm~\cite{ghorbani2019towards}. It first divides images into superpixels using SLIC algorithm~\cite{achanta2012slic} with three resolutions of 15, 50, and 80 segments for each image. Next, it generates representations of these superpixels as an output of the \textit{mixed4c} layer of GoogLeNet~\cite{szegedy2015going} trained on the ImageNet dataset. Then, for each class separately, corresponding representations are clustered using the k-means algorithm with $k=25$ and filtered to remove infrequent and unpopular clusters (as described in~\cite{ghorbani2019towards}). This results in around 18 concepts per class and approximately $18,000$ concepts for the whole ImageNet dataset. They could be directly used as visual words. However, such words would be exclusive for particular classes, and some of them would be ambiguous due to the small TCAV score~\cite{kim2018interpretability}. Hence, to obtain a reliable dictionary with visual words shared between classes, we filter out $12,000$ concepts with the smallest TCAV score and cluster the remaining $6,000$ concepts using the k-means algorithm into 50 clusters treated as visual words (see Figures~\ref{fig:wc} and~\ref{fig:best_concepts_mini}). We do not treat the number of clusters as a tunable hyperparameter. Instead, we set a fixed number of clusters, ensuring various concepts and making user studies feasible.

\subsection{Generating a self-supervised representation}
We examine four self-supervised methods: MoCo~v1~\cite{he2020momentum}, SimCLR~v2~\cite{chen2020big}, BYOL~\cite{grill2020bootstrap}, and SwAV~\cite{caron2020unsupervised}. For all of them, we use publicly available models trained on ImageNet\footnote{We use the following implementations of self-supervised methods: https://github.com/\{google-research/simclr, yaox12/BYOL-PyTorch, facebookresearch/swav, facebookresearch/moco\}. We use ResNet-50~(1x) variant for each self-supervised method.}. Although they all use the penultimate layer of ResNet-50 to generate representations, their training hyperparameters differ, which is presented in Table~\ref{tab:ss_train_details}.

\subsection{Assigning visual words}
To assign a superpixel to a visual word, we first pass it through the GoogLeNet to generate a representation from the \textit{mixed4c} layer (similarly to generating visual words). Since all concepts considered in Section~\ref{subsec:generate_visual_words} are grouped into 50 clusters (visual words), we use a two-stage assignment. First, we find the closest concept and then assign the superpixel to the visual word containing this concept.

\subsection{Training probing tasks}
We use a logistic regression classifier with the LBFGS solver~\cite{lbfgs} to train all diagnostic classifiers. As input, we use representations generated by the self-supervised methods. The output depends on the probing task.
In the case of \textit{Word Content}, we train 50 classifiers corresponding to 50 visual words. Furthermore, we expect an image to be assigned to a particular visual word if at least one of its superpixels is assigned to it. Finally, we report the average AUC scores over 50 classifiers (see Table~\ref{tab:results}).
To formulate a classification setup in the \textit{Sentence Length} probing task, we group the possible output into six equally-sized bins (see Table~\ref{tab:pt_details}), resulting in one-vs-one OVO AUC, which is resistant to class imbalance.
A similar procedure is applied to the \textit{Character Bin} probing tasks.
SOMO is formulated as a binary classification task in which we predict whether the image was modified or not. We train two separate classifiers for two use-cases, SOMO far and SOMO close with balanced training and validation sets.

We conduct all of our experiments on the ImageNet dataset~\cite{deng2009imagenet} with  standard train/validation split. Moreover, we apply random over-sampling if needed to deal with the imbalanced classes.

\begin{table}[t]
    \centering
    \caption{Bins ranges corresponding to the classes in \textit{Sentence Length} (SL) and \textit{Character Bin} (CB) probing tasks.}
    \begin{tabular}{c | c | c | c} \hline
    bin & SL & CB shape & CB color \\
    \hline
    0 &  $<18$  &  $<0.153$ & $<0.063$\\
    1 &  $[18, 21)$ & $[0.153, 0.207)$ & $[0.063, 0.085)$ \\
    2 &  $[21, 23)$ & $[0.207, 0.263)$& $[0.085, 0.104)$\\
    3 &  $[23, 26)$ & $[0.263, 0.336)$& $[0.104, 0.125)$\\
    4 &  $[26, 28)$ & $[0.336, 0.462)$&$[0.125, 0.155)$\\
    5 &  $28\leq$  & $0.462\leq$ & $0.155\leq$\\
    \hline
    \end{tabular}
    \label{tab:pt_details}
\end{table}

\section{User studies}
\label{sec:user_studies}

While the cognitive visual systematic introduced in Section~\ref{subsec:marr} presents the possible way of obtaining the meaning of visual words, it requires human observers to reliably decide which visual features should be assigned to particular visual words. Hence, in this section, we describe user studies conducted to establish this assignment.

Overall, 40 volunteers participated in the study (30 males and 10 females aged 29 $\pm$ 10 years) recruited online. $62.5 \% $ of the participants were students/graduates of computer science and related fields, and the remaining attendees represented various backgrounds.

\begin{figure}[h!]
\centering
\includegraphics[width=0.8\linewidth]{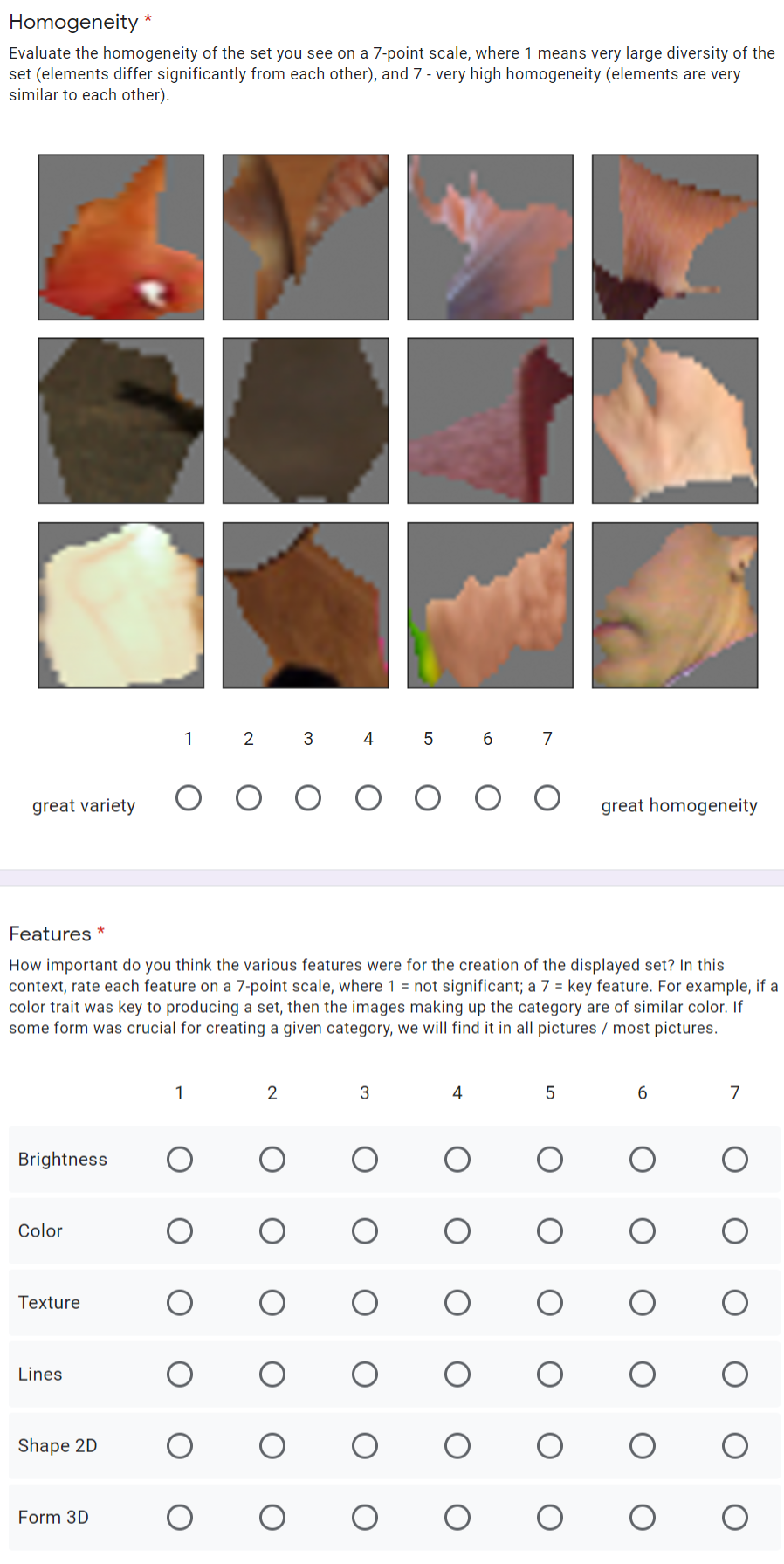}
\caption{Sample question from our user study that allows in-depth analysis of the Word Content probing task using Marr's computational theory of vision.}
\label{fig:user_studies_print_screen}
\end{figure}

Users completed an online questionnaire. Their task was to assess the similarity of superpixels representing a visual word and provide key features associated with this visual word. To this end, users were presented with 20 visual words consisting of 12 representative superpixels (close to the visual word center) each. Participants were instructed to use Likert scales with seven numerical responses with only endpoints labeled (1 and 7) for clarity. First, they were asked to evaluate the homogeneity of a given set (scale endpoints: great variety; great homogeneity; see Figure~\ref{fig:user_studies_print_screen}). Next, they evaluated to what extent a given feature was essential for visual word creation. In reference to Marr’s computational theory of vision \cite{Marr:1982:VCI:1095712} (see Section~\ref{subsec:marr}), six features were taken into consideration: light intensity (brightness), color, texture, lines, shape (Marr's 2.5D sketch) and form (Marr's 3D model representation). Scale endpoints were labeled as a not significant feature and a key feature (see Figure~\ref{fig:user_studies_print_screen}).

Before the main task, users obtained an instruction that included sample visual words with particular features (selected by a cognitivist). They also underwent two training trials in order to be familiarized with the task. There were no time constraints for trial or task completion. The order of visual words and on-screen localization of superpixels were semi-randomized for each participant.

Due to the high number of visual words, the assessment of all 50 visual words would be tedious for the users. That is why we decided to limit our user study to the twenty most reliable visual words. They were chosen based on the results of \textit{Word Content} probing task by selecting best and worst-performing clusters, as well as the ones with the largest performance difference between considered self-supervised models.

Based on the results of the user studies, we select the most representative visual words for each of the six features: brightness, color, texture, lines, shape, and form. Those words are then used to obtain detailed results of the \textit{Word Content} probing task presented in Table~\ref{tab:results_wc_user}.

\section{Results and Discussion}
\label{sec:results}

As we show in Table~\ref{tab:results}, all self-supervised representations retain information about semantic knowledge, complexity, and image consistency. However, SimCLR~v2 surpasses other methods in all probing tasks, except CB color. Moreover, the performance on probing tasks does not correlate with the accuracy of the target task. In the following, we analyze those aspects in greater detail.

\begin{table*}[h]

\caption{\label{tab:results}AUC score for our probing tasks (WC, MWC, SL, CB, and SOMO) and accuracy on the linear evaluation (Target) for the considered self-supervised methods. Like the linear evaluation, our probing tasks are also trained on top of the frozen backbone network, and test accuracy is used to analyze representation properties. Hence, they provide complementary knowledge about the representation.}
\begin{center}

\begin{tabular}{l|c||c|c|c|c|c|c|c} 
\hline
& \textbf{Target} & \multicolumn{7}{c}{\textbf{Probing tasks (ours)}}\\
& & \textbf{WC} & \textbf{MWC} & \textbf{SL} & \textbf{CB} \textbf{shape} & \textbf{CB color} & \textbf{SOMO far}& \textbf{SOMO close} \\
\hline
MoCo v1 & $0.606$ & $0.793$ & $0.763$ & $0.771$ & $0.797$ & $0.872$ & $0.850$ & $0.830$ \\
SimCLR v2 & $0.717$ & $\mathbf{0.811}$ & $\mathbf{0.777}$ & $\mathbf{0.775}$ & $\mathbf{0.850}$ & $0.876$ & $\mathbf{0.878}$ & $\mathbf{0.857}$ \\
BYOL & $ 0.723$ & $0.803$ & $0.775$ & $0.770$ & $0.844$ & $\mathbf{0.893}$ & $0.845$ & $0.817$\\
SwAV & $\mathbf{0.753}$ & $0.802$ & $0.776$ & $0.769$ & $0.842$ & $0.879$ & $0.856$ & $0.839$\\
\hline
\end{tabular}%
\end{center}

\end{table*}

\subsection{Self-supervised representations contain semantic knowledge which does not correlate with the target task}
As reported in Table~\ref{tab:results}, the AUC scores for \textit{Word Content} probing task vary from $0.793$ for MoCo~v1 to $0.811$ for SimCLR~v2. This shows that considered self-supervised methods can predict which visual words are present in the image, i.e. they code the semantic knowledge in the generated representations.

Surprisingly, although the examined self-supervised methods diverse in target task accuracy, they all have a similar level of semantic knowledge. For instance, MoCo~v1 obtains the worst target task accuracy (60.6\%), but its results for the WC probing task are on par with more accurate self-supervised methods. Moreover, although SwAV has the highest accuracy on the target task, it does not provide the best performance in terms of semantic knowledge. This finding supports the conclusion from~\cite{geirhos2020surprising} that semantic knowledge only partially contributes to the target task accuracy.

\subsection{Certain types of visual words are represented better than the others, depending on the method}
\begin{figure}[t!]
     \centering
     \begin{subfigure}[b]{0.9\linewidth}
         \centering
         \includegraphics[width=0.9\textwidth]{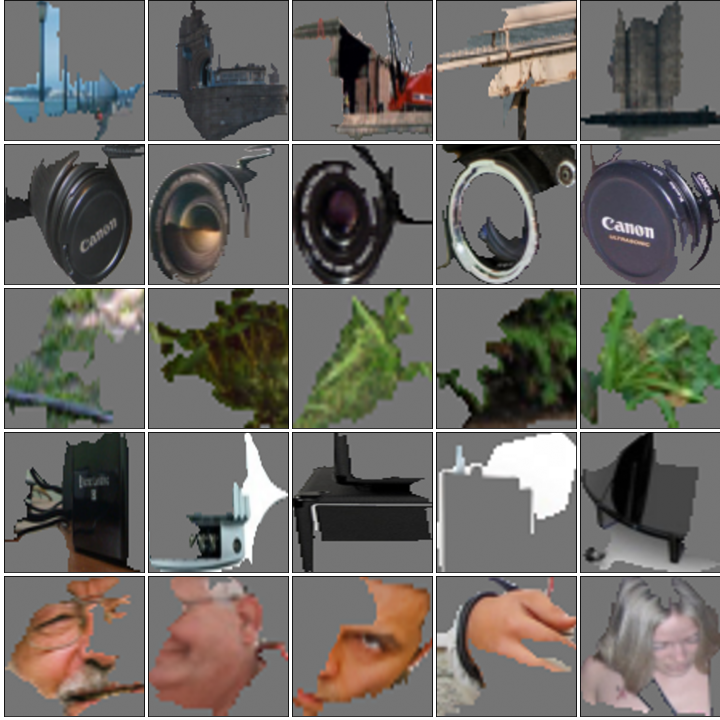}
         \caption{Best predicted visual words.}
     \end{subfigure}\hfill%
     \begin{subfigure}[b]{0.9\linewidth}
         \centering
         \includegraphics[width=0.9\textwidth]{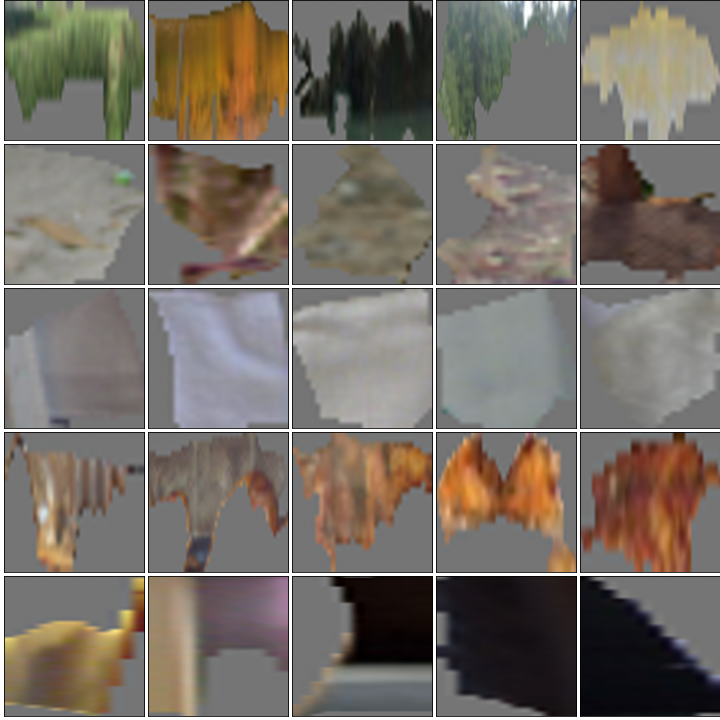}
         \caption{Worst predicted visual words.}
     \end{subfigure}
     \caption{Visualization of the best (a) and the worst (b) predicted visual words according to the results of the WC probing task. It supports the results from Table~\ref{tab:results_wc_user} that self-supervised representations contain more information about lines and forms than about textures.} 
    \label{fig:best_concepts_mini}
\end{figure}

According to the results presented in Table~\ref{tab:results_wc_user} and Figure~\ref{fig:best_concepts_mini}, self-supervised representations have more knowledge about visual words containing forms and lines than about those containing shapes and textures. This may indicate that the representations are lines- and form-biased, which sheds new light on this problem, considering that according to~\cite{geirhos2020surprising}, self-supervised representations are texture-biased.
Moreover, the information encoded by various self-supervised methods differs. It is especially visible for brightness and color, where the MoCo~v1 works significantly worse than the remaining methods. We assume that it is caused by the lack of projection head in the former, which is important due to loss of information induced by the contrastive loss~\cite{chen2020simple}.

\begin{table*}[h]
    \caption{\label{tab:results_wc_user}Detailed AUC scores for our WC probing tasks. The results are presented for six visual concepts from Marr’s computational theory of vision. The colors denote higher (orange) or lower (blue) AUC score compared to the overall performance obtained for all visual words and demonstrate the biases in the self-supervised representations.}
    \begin{center}
    \begin{tabular}{l|c|c|c|c|c|c|c} 
    \hline
    & 
    & \multicolumn{6}{c}{\textbf{Types of visual words}}\\
    & all visual words & \textbf{brightness} & \textbf{color} & \textbf{texture} & \textbf{lines} & \textbf{shape} & \textbf{form} \\
    \hline
    MoCo v1 & $0.793$ & \cellcolor{blue2}{$0.777$} & \cellcolor{blue3}{$0.769$} & \cellcolor{blue1}{$0.785$} & \cellcolor{red6}{$0.847$} & \cellcolor{blue1}{$0.784$} & \cellcolor{red5}{$0.836$} \\
    SimCLR v2 & $0.811$ & \cellcolor{red2}{$0.831$} & \cellcolor{red3}{$0.832$} & \cellcolor{blue1}{$0.804$} & \cellcolor{red5}{$0.852$} & \cellcolor{blue1}{$0.810$} & \cellcolor{red5}{$0.854$} \\
    BYOL & $0.803$ & \cellcolor{red1}{$0.809$} & \cellcolor{red1}{$0.807$} & \cellcolor{blue1}{$0.795$} & \cellcolor{red5}{$0.852$} & \cellcolor{blue1}{$0.798$} & \cellcolor{red5}{$0.848$} \\
    SwAV & $0.802$ & \cellcolor{red2}{$0.819$} & \cellcolor{red2}{$0.820$} & \cellcolor{blue1}{$0.796$} & \cellcolor{red5}{$0.851$} & \cellcolor{grey}{$0.802$} & \cellcolor{red5}{$0.849$} \\
    \hline
    \end{tabular}
    \end{center}
    
\end{table*}

\subsection{The same visual word in a pair of images usually brings their representations closer}
\begin{figure}[t!]
     \centering
     \begin{subfigure}{0.45\textwidth}
         \centering
         \includegraphics[width=\textwidth]{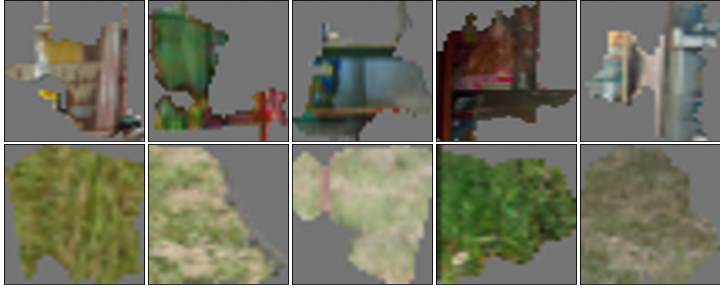}
         \caption{Attracting representations.}
         \label{subfig:mwc_closer}
     \end{subfigure}\hfill%
     \begin{subfigure}{0.45\textwidth}
         \centering
         \includegraphics[width=\textwidth]{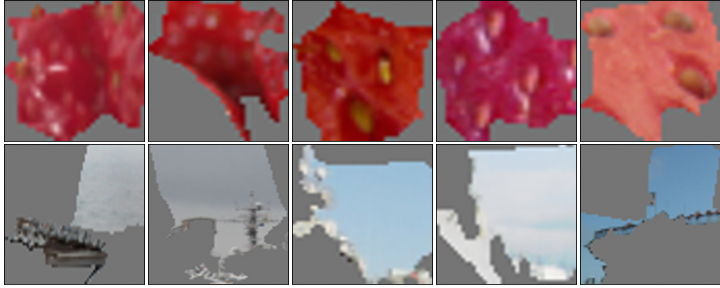}
         \caption{Neutral for representations.}
     \end{subfigure}
     \hfill
     \begin{subfigure}{0.45\textwidth}
         \centering
         \includegraphics[width=\textwidth]{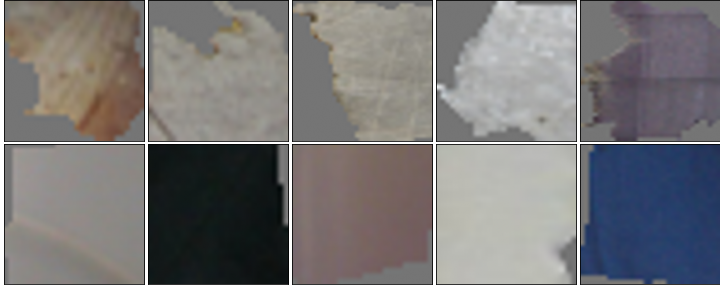}
         \caption{Pushing representations away.}
         \label{subfig:mwc_farther}
     \end{subfigure}
     \caption{Sample visual words that attract (a), are neutral (b), or push away (c) the representations of two images. One can observe that the visual words that attract representations include words with complicated forms or green areas. On the other hand, visual words that push representations away contain fine textures.}
    \label{fig:best_mwc}
\end{figure}

The results of the MWC probing task presented in Table~\ref{tab:results_mwc_user} show that the same visual word in a pair of images usually brings their representations closer. This is true for almost all visual words (45 out of 50), and especially for those presented in Figure~\ref{subfig:mwc_closer} that contain complicated forms and lines or green areas. The remaining 5 visual words, usually corresponding to fine textures (see Figure~\ref{subfig:mwc_farther}), are neutral or pushing representations away.

\begin{table*}[h]
\caption{The attraction coefficient depending on the type of visual word. The colors represent whether a given type of visual word attracts (orange) or pushes away (blue) a pair of representations. In most cases, one can observe that the same visual words in a pair of images bring their representations closer.}
    
    \label{tab:results_mwc_user}
    \begin{center}
    \begin{tabular}{l|c|c|c|c|c|c} 
    \hline
    & \multicolumn{6}{c}{\textbf{Types of visual words}}\\
    & \textbf{brightness} & \textbf{color} & \textbf{texture} & \textbf{lines} & \textbf{shape} & \textbf{form} \\
    \hline
    MoCo v1 & \cellcolor{red4}$0.872$ & \cellcolor{red4}$0.875$ & \cellcolor{red4}$0.937$ & \cellcolor{red6}$1.839$ & \cellcolor{red3}$0.677$ & \cellcolor{red6}$1.928$ \\
    SimCLR v2 & \cellcolor{red2}$0.409$ & \cellcolor{red2}$0.424$ & \cellcolor{red2}$0.442$ & \cellcolor{red4}$0.803$ & \cellcolor{red2}$0.500$ & \cellcolor{red3}$0.593$ \\
    BYOL & \cellcolor{red2}$0.396$ & \cellcolor{red2}$0.379$ & \cellcolor{red3}$0.502$ & \cellcolor{red2}$0.336$ & \cellcolor{grey}$-0.007$ & \cellcolor{red3}$0.566$ \\
    SwAV & \cellcolor{red2}$0.382$ & \cellcolor{red2}$0.445$ & \cellcolor{red1}$0.150$ & \cellcolor{blue1}{$-0.119$} & \cellcolor{blue1}$-0.118$ & \cellcolor{red2}$0.248$ \\
    \hline
    \end{tabular}
    \end{center}
\end{table*}

Interestingly, SwAV differs from the other methods in the case of lines and shapes, and BYOL differs in the case of shape. As in both cases, the presence of those features usually does not bring representations learned by those methods closer together.

\subsection{Self-supervised representations contain information about semantic complexity that differs between methods}
\begin{figure*}
\begin{subfigure}{0.9\textwidth}
\centering
\includegraphics[width=\linewidth]{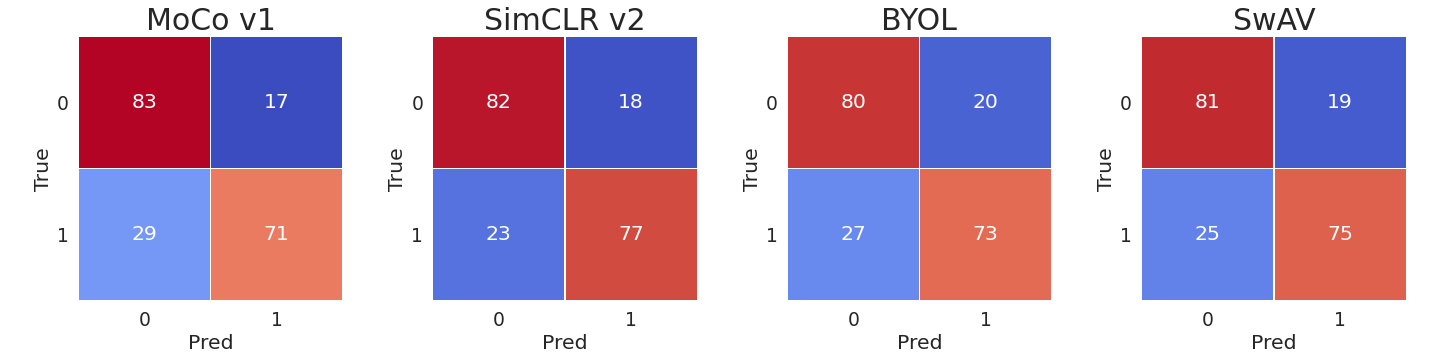}
\caption{SOMO.}
\end{subfigure}\hfill%
\begin{subfigure}{0.9\textwidth}
\centering
\includegraphics[width=\linewidth]{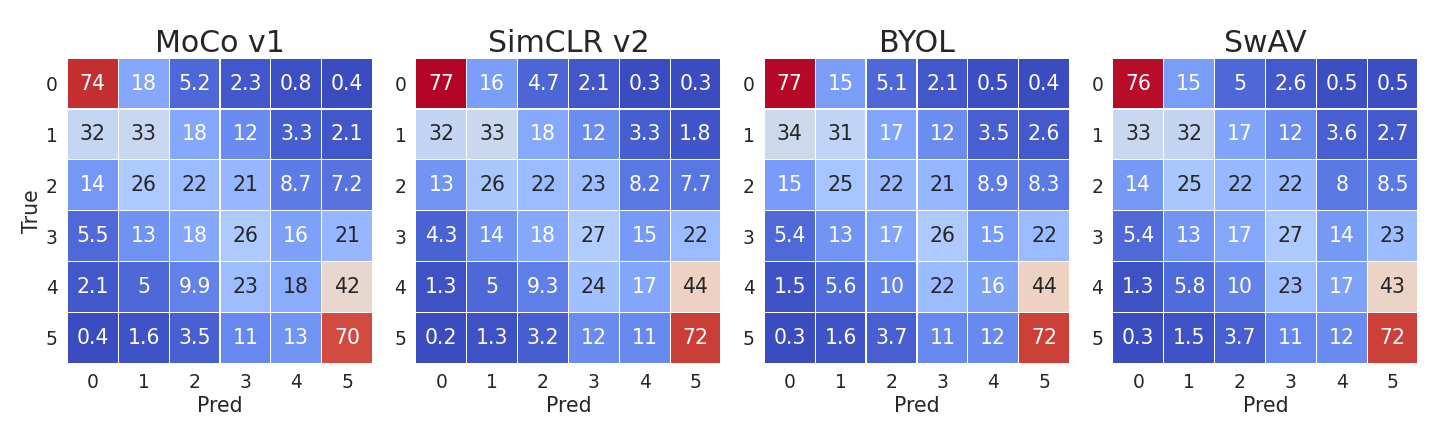}
\caption{Sentence length.}
\end{subfigure}\hfill%
\begin{subfigure}{0.9\textwidth}
\centering
\includegraphics[width=\linewidth]{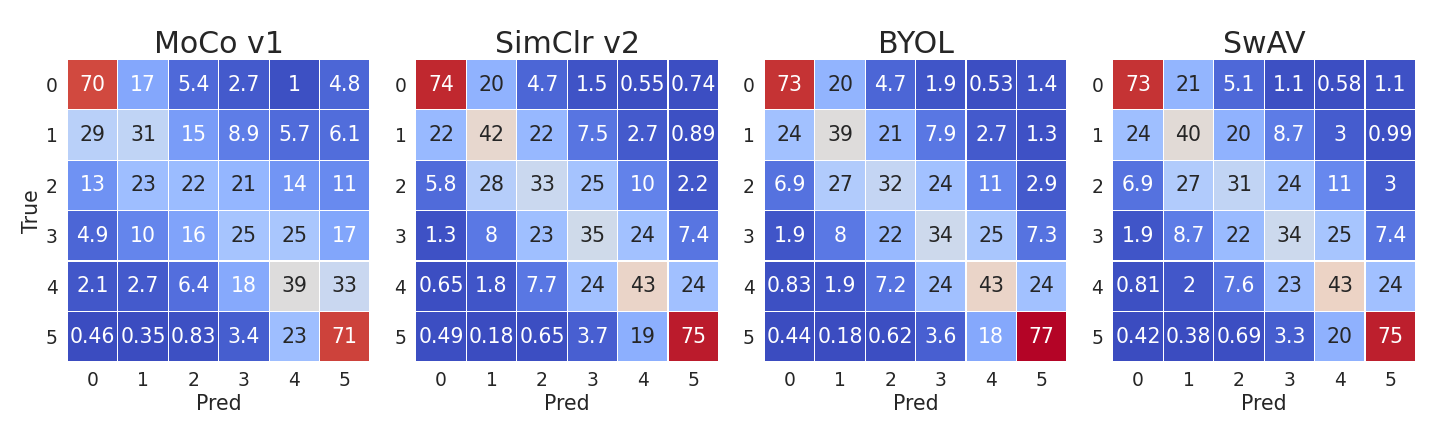}
\caption{\textit{Character Bin} shape.}
\end{subfigure}\hfill%
\begin{subfigure}{0.9\textwidth}
\centering
\includegraphics[width=\linewidth]{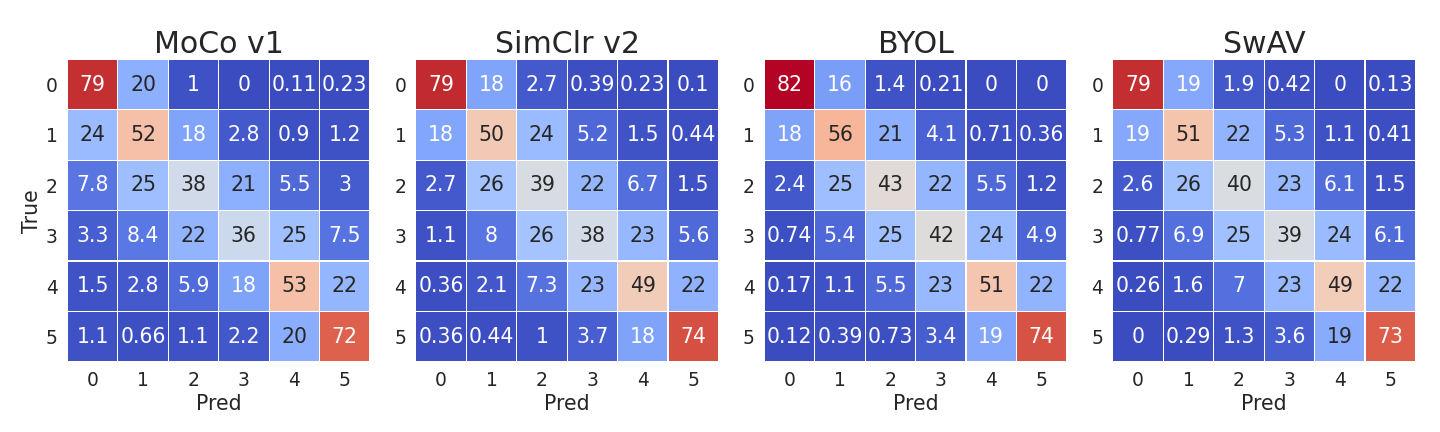}
\caption{\textit{Character Bin} color.}
\end{subfigure}
\caption{Confusion matrices for \textit{Sentence Length} and \textit{Character Bin} probings (results in \%). The results indicate that the ability of self-supervised representations to retain information about complexity differs depending on the level of image complexity. Moreover, even though the final AUC of SL and CB for self-supervised methods are similar, their confusion matrices differ.}
\label{fig:cm}
\end{figure*}

Based on the results in Table~\ref{tab:results} and Figure~\ref{fig:cm}, we observe that considered self-supervised methods code the level of semantic complexity, as they all obtain approximately 0.77 AUC for \textit{Sentence Length}, and even higher AUCs are observed for CB shape and color (from 0.797 to 0.893 AUC). Moreover, when it comes to recognizing variance in superpixel color, BYOL works best, in contrast to all other probing tasks, where SimCLR~v2 has the highest AUC. The potential reason for this behavior is the fact that a positive pair with similar color histograms provide more information in BYOL than in SimCLR, as presented in Section~5 of~\cite{grill2020bootstrap}. Therefore, BYOL puts more attention on the color characteristic.

\subsection{Self-supervised representations contain information about semantic consistency that differs between methods}
\begin{figure}[h]
\centering
\begin{subfigure}{0.45\textwidth}
\centering
\includegraphics[width=\linewidth]{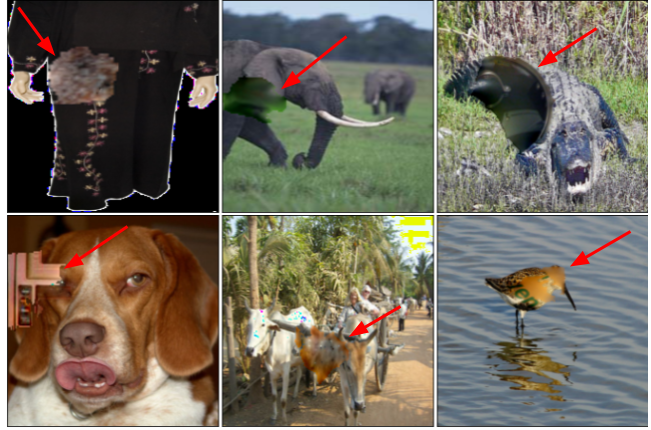}
\caption{Correctly classified.}
\end{subfigure}\hfill%
\begin{subfigure}{0.45\textwidth}
\centering
\includegraphics[width=\linewidth]{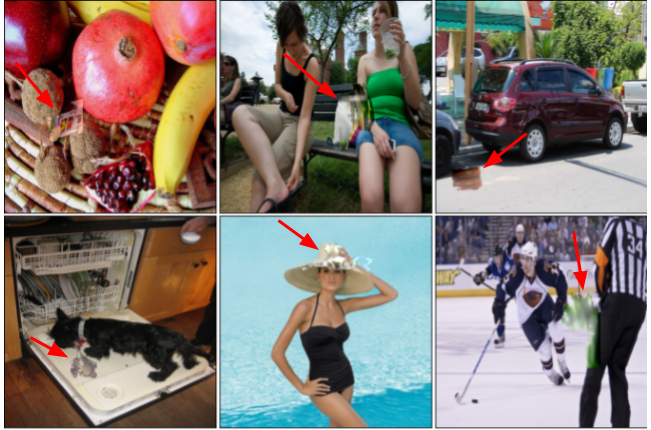}
\caption{Incorrectly classified.}
\end{subfigure}
\caption{Sample images from the SOMO probing task, correctly classified by all methods (a) and by none of them (b). One can observe that probing struggles with more subtle changes, which are still visible to the human eye. Notice that red arrows indicate alterations to the image.}
\label{fig:pos_neg_arrows}
\end{figure}

\begin{figure}[h!]
\centering
\begin{subfigure}{0.45\textwidth}
\centering
\includegraphics[width=\linewidth]{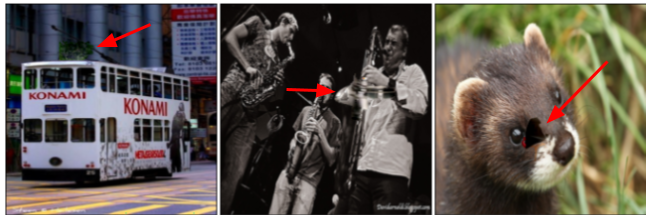}
\caption{Correctly classified only by MoCo v1.}
\end{subfigure}\hfill%
\begin{subfigure}{0.45\textwidth}
\centering
\includegraphics[width=\linewidth]{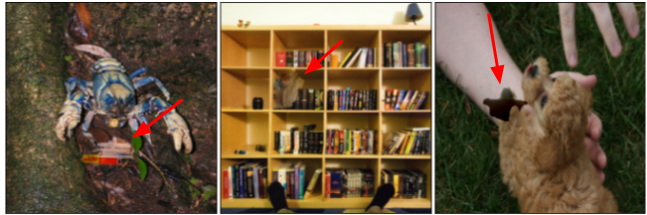}
\caption{Correctly classified only by SimCLR v2.}
\end{subfigure}\hfill%
\begin{subfigure}{0.45\textwidth}
\centering
\includegraphics[width=\linewidth]{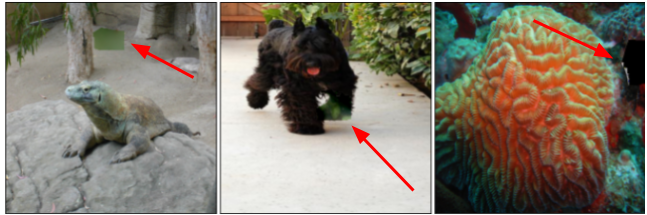}
\caption{Correctly classified only by BYOL.}
\end{subfigure}\hfill%
\begin{subfigure}{0.45\textwidth}
\centering
\includegraphics[width=\linewidth]{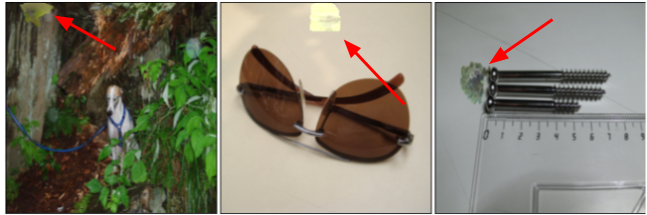}
\caption{Correctly classified only by SwAV.}
\end{subfigure}
\caption{Sample images from SOMO probing task, correctly classified when embedded by MoCov1 (a), SimCLR v2 (b), BYOL (c), or SwAV (d) only. One can observe no clear differences between the types of inconsistencies classified correctly and incorrectly by the particular methods. Notice that red arrows indicate alterations to the image.}
\label{fig:method_analysis_arrows}
\end{figure}

The results of the SOMO probing task in Table~\ref{tab:results} and Figure~\ref{fig:cm} show that self-supervised representations reflect changes in the center of the image. However, as presented in Figure~\ref{fig:pos_neg_arrows}, the probing classifier struggles with more subtle changes, which are still visible to the human eye. Moreover, SimCLR~v2 has the highest ability to recognize altered images, but surprisingly, BYOL has the lowest performance. However, as shown in Figure~\ref{fig:pos_neg_arrows} and~\ref{fig:method_analysis_arrows}, there are no visible reasons for this result.
Overall, our results are in line with~\cite{hendrycks2019using}, which claims that self-supervised methods improve out-of-distribution detection. However, they are in contradiction to our previous results~\cite{visual_probing_ijcai}, where the replaced superpixel is selected entirely randomly (without center bias). Nevertheless, we decided to change replacement to center-biased in this work because they better correspond to semantic inconsistency.

\subsection{The ability to distinguish altered images depends on how often the removed visual word co-occurs with the replacement} 
As presented in Table~\ref{tab:results}, SOMO far has higher performance than SOMO close. It is expected because recognizing alterations obtained by replacing a visual word with a non-fitting one is simpler. However, this difference in performance for all self-supervised representations leads us to believe that there is a family of alterations that might not be reflected well enough in a self-supervised representation. Hence, considering that even minor alterations might lead to a change in the prediction~\cite{43405}, this disability might pose a risk to the stability of the classification results.

\section{Conclusions}

In this work, we introduce a novel visual probing framework that analyzes the information stored in self-supervised image representations. It is inspired by probing tasks employed in NLP and requires similar taxonomy. Hence, we propose a set of intuitive mappings between visual and textual modalities to construct visual sentences, words, and characters. Moreover, we provide a cognitive visual systematic that identifies a visual word with structural features from Marr's computational theory~\cite{Marr:1982:VCI:1095712} and provide the meaning of the words.

The results of the provided experiments confirm the effectiveness and applicability of this framework in understanding self-supervised representations. We verify that the representations contain information about semantic knowledge, complexity, and consistency of the images.  Moreover, a detailed analysis of each probing task reveals differences in the representations encoded by various methods, providing complementary knowledge to the accuracy of linear evaluation.

Finally, we show that the relations between language and vision can serve as an effective yet intuitive tool for explainable AI. Hence, we believe that our work will open new research directions in this domain.

\begin{sidewaystable}
    \centering
    \caption{{Differences between architecture and training of the considered self-supervised methods.}}
    \begin{tabularx}{\linewidth}{p{2em}|X|X|X|X|X}
    \toprule
    \multicolumn{2}{c|}{}  &  MoCo v1 & SimCLR~v2 & BYOL & SwAV \\
    \hline
    \multirow{7}{*}{\begin{sideways}Architecture\end{sideways}}\\
    & InfoNCE & yes & yes & no & no \\ 
    & Positive pairs & yes & yes & yes & yes \\  
    & Negative pairs & yes, minibatches queue & yes, large batches & no & no \\  
    & Online to target network & copied with momentum & same & copied with momentum & same \\
    & Size of patches & 224x224 & 224x224 & 224x224 & 224x114 and 96x96 \\
    & Augmentations & resize, crop, color jittering, horizontal flip, grayscale conv.& crop, resize, horizontal flip, color distortion, grayscale conv., Gaussian blur, solarization & like in SimCLR~v2 & two types of crops, small and original, the rest like in SimCLR~v2\\
    & Projection & no & yes & yes & yes\\
    \midrule
    \multirow{3}{*}{\begin{sideways}Training\end{sideways}}\\
    & epochs & 200 & 600 & 300 & 800 \\
    & Batch size & 256 & 2048 & 4096 & 4096 \\
    & Time of training & 53 & 170 & not mentioned & 49\\
    \bottomrule
    \end{tabularx}
    \label{tab:ss_train_details}
\end{sidewaystable}

\begin{IEEEbiography}[
{
\includegraphics[width=1in,height=1.25in,clip,keepaspectratio]{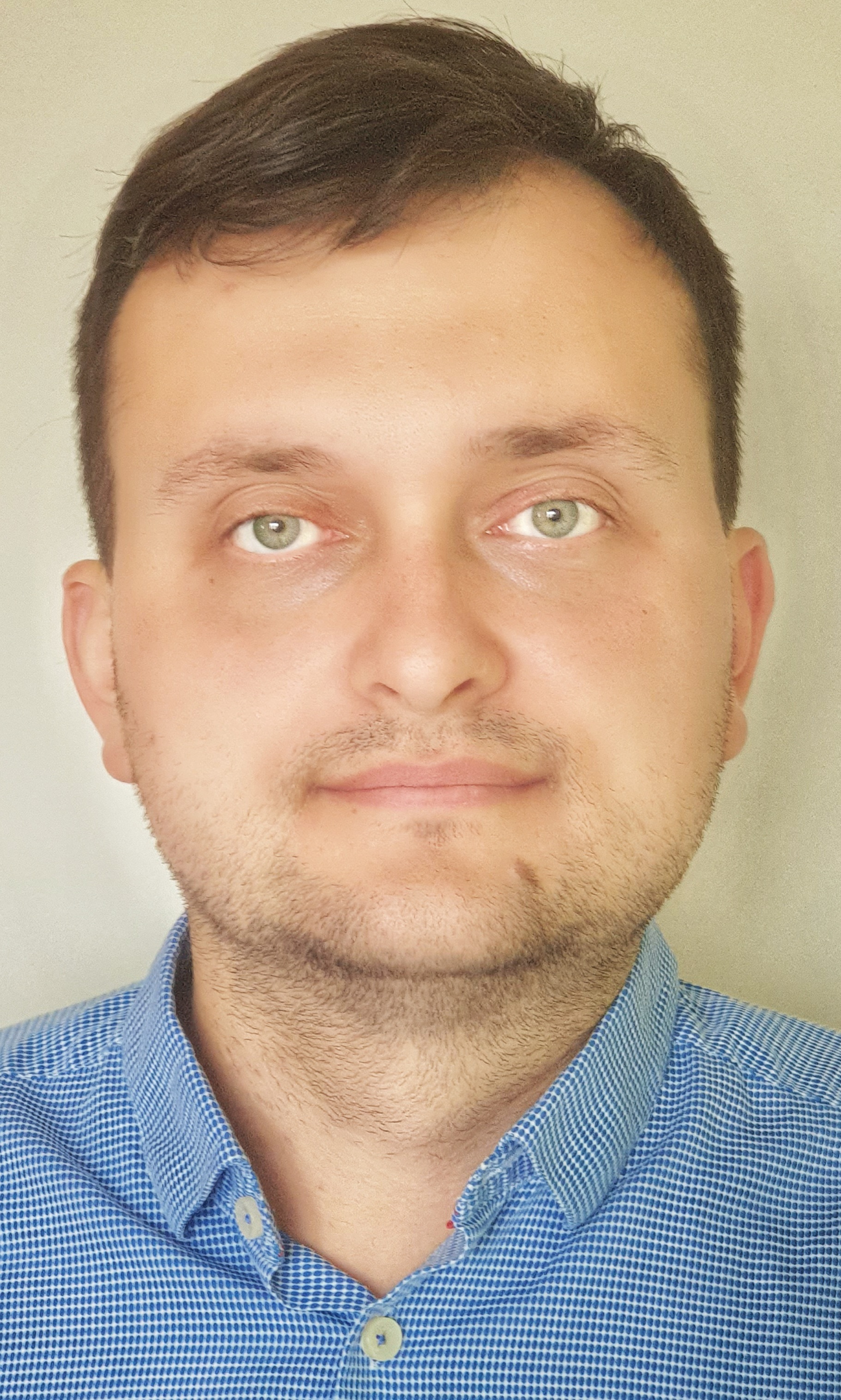}
}
]{Witold Oleszkiewicz}
is an Assistant and a Ph.D. student in the Division of Artificial Intelligence in the Institute of Computer Science at Warsaw University of Technology. He obtained his M.Sc. in Computer Science from the Institute of Computer Science at Warsaw University of Technology in 2017. His professional appointments include work with Samsung in 2013 and Braster from 2015 to 2018, where he worked on the use of machine learning in breast cancer detection. He was a Visiting Scholar at Stanford University in 2018, where he worked on privacy-preserving generative models, and at New York University in 2019, where he worked on understanding the robustness of deep learning for breast cancer screening.
\end{IEEEbiography}
\begin{IEEEbiography}[
{
\includegraphics[width=1in,height=1.25in,clip,keepaspectratio]{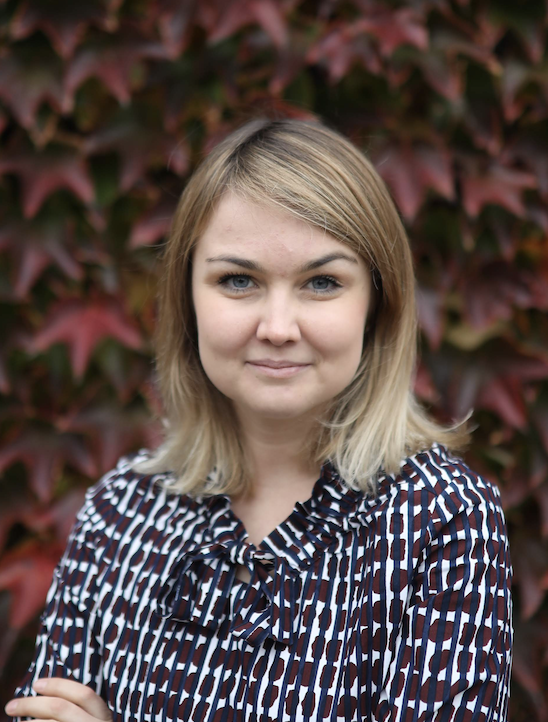}
}
]{Dominika Basaj}
is a Senior AI Engineer in Tooploox. She obtained her Master’s degree in Quantitative Methods in Economics and Information Systems at the Warsaw School of Economics in 2016. She was developing machine learning models in financial institutions. Her research focuses on the interpretability and robustness of neural networks. In 2019 she was a Visiting Researcher at the Nanyang University of Technology, where she worked on discourse-aware neural machine translation, and at the University of California at Davis, where she worked on the prediction of protein structure.
\end{IEEEbiography}
\begin{IEEEbiography}[
{
\includegraphics[width=1in,height=1.25in,clip,keepaspectratio]{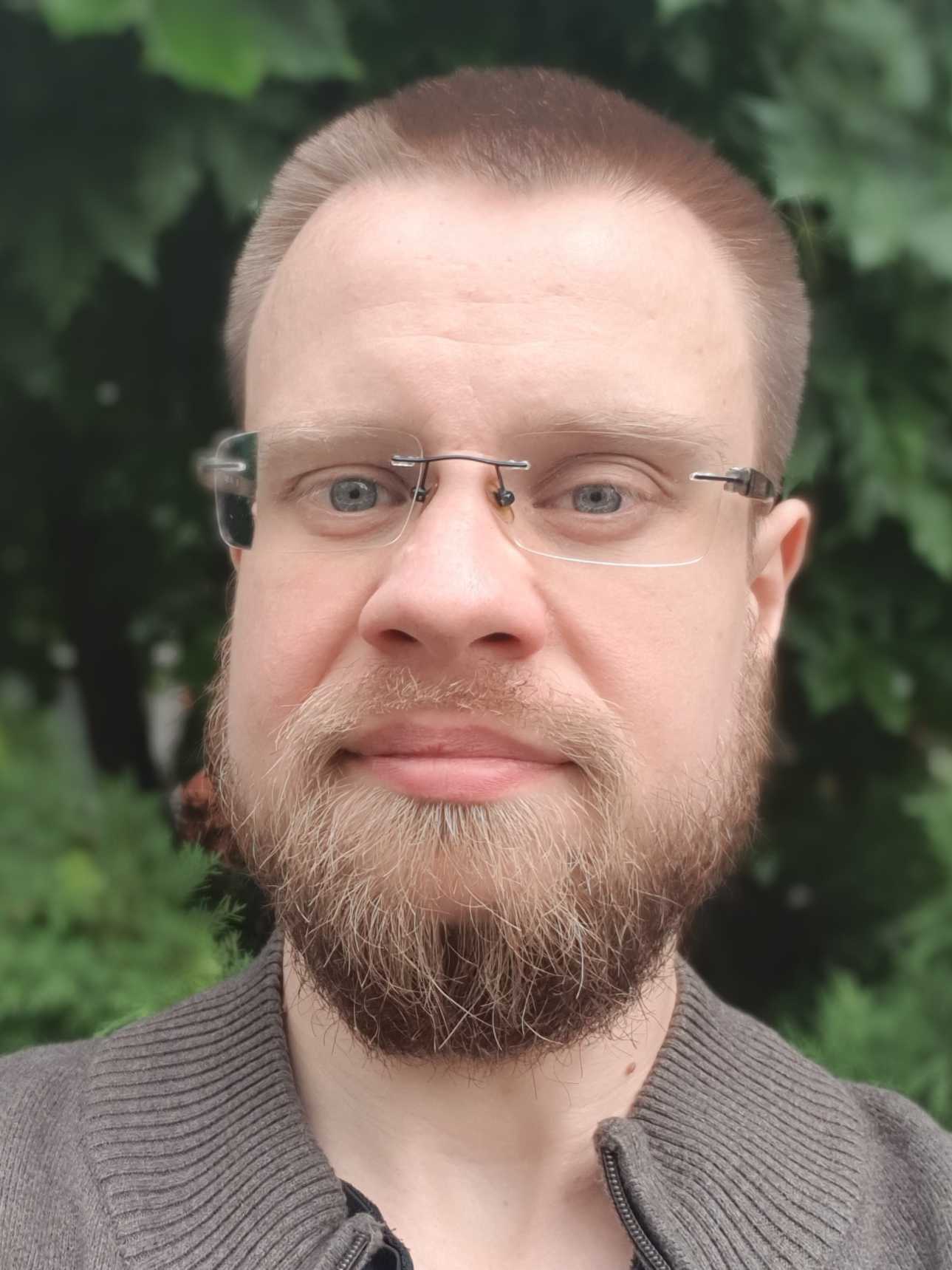}
}
]{Igor Sieradzki}
is an Assistant and a Ph.D. student in the Faculty of Mathematics and Computer Science in the Institute of Computer Science and Computer Mathematics at Jagiellonian University in Kraków since 2019. He obtained his M.Sc in Computer Science on Active Learning in computer-aided drug design from Jagiellonian University in 2016. Before the academic position, he worked with Applica.ai on the modern use of deep learning in natural language processing. His research internships include a stay at the University of Edinburgh in 2015.
\end{IEEEbiography}
\begin{IEEEbiography}[
{
\includegraphics[width=1in,height=1.25in,clip,keepaspectratio]{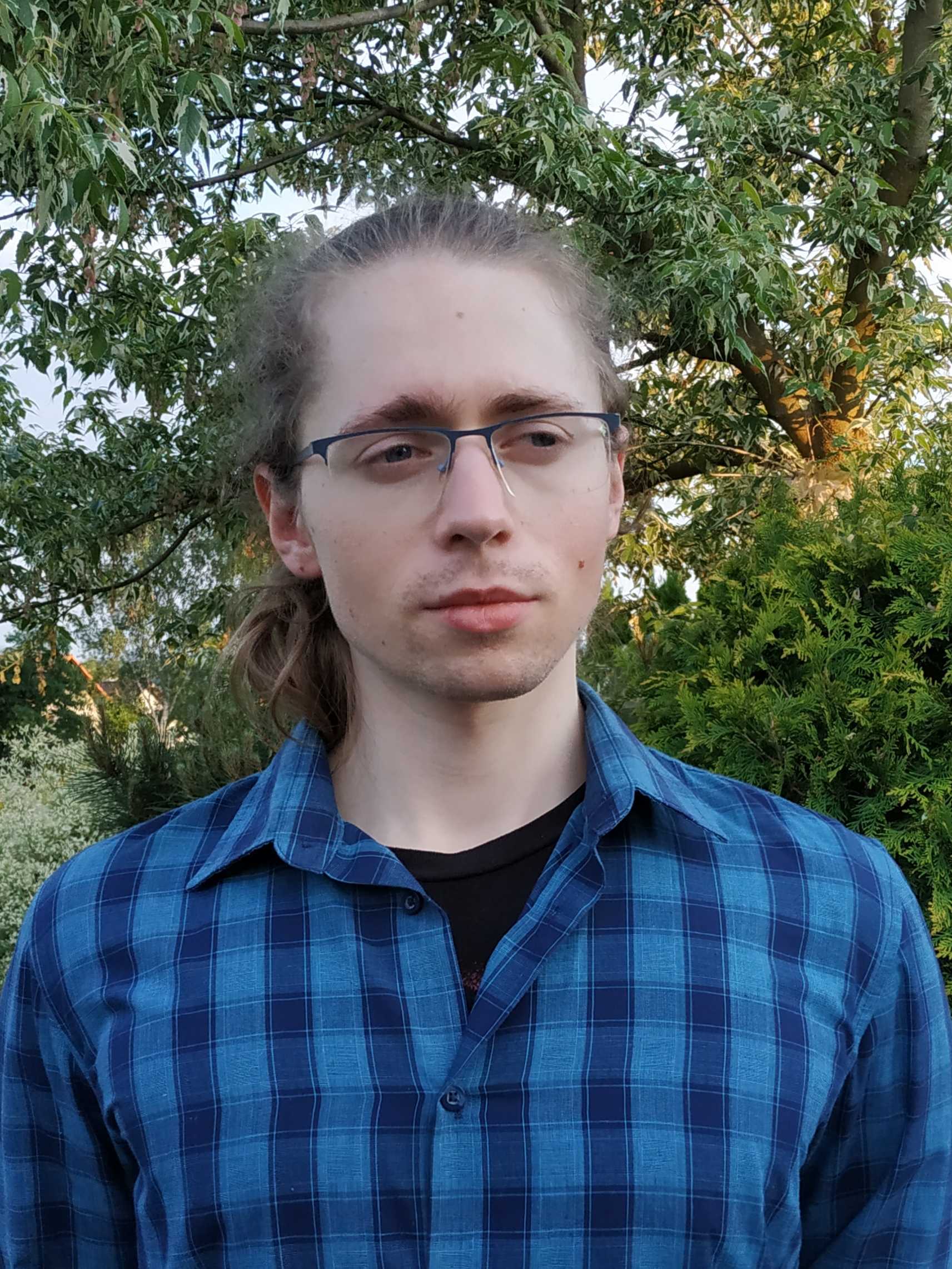}
}
]{Michał Górszczak}
is a Master's student in the Faculty of Mathematics and Computer Science at Jagiellonian University in Kraków. He obtained his B.Eng. in Applied Computer Science from the University of Science and Technology in Kraków in 2019.
\end{IEEEbiography}

\begin{IEEEbiography}[
{
\includegraphics[width=1in,height=1.25in,clip,keepaspectratio]{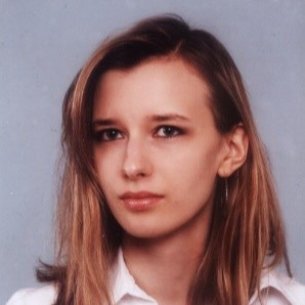}
}
]{Barbara Rychalska}
is a Ph.D. student at the Faculty of Mathematics and Information Science at the Warsaw University of Technology. She obtained her Master's degree in Computer Science at the Warsaw University of Technology in 2016, having also studied applied linguistics at the Warsaw University. She is an AI Research Scientist at Synerise, where she works on topics ranging from natural language processing to recommender systems. Previously, she worked at Samsung R\&D Research Institute Warsaw and Findwise AB as an AI researcher. She was a Visiting Scientist at the Nanyang Technological University in Singapore in 2019.
\end{IEEEbiography}
\begin{IEEEbiography}[
{
\includegraphics[width=1in,height=1.25in,clip,keepaspectratio]{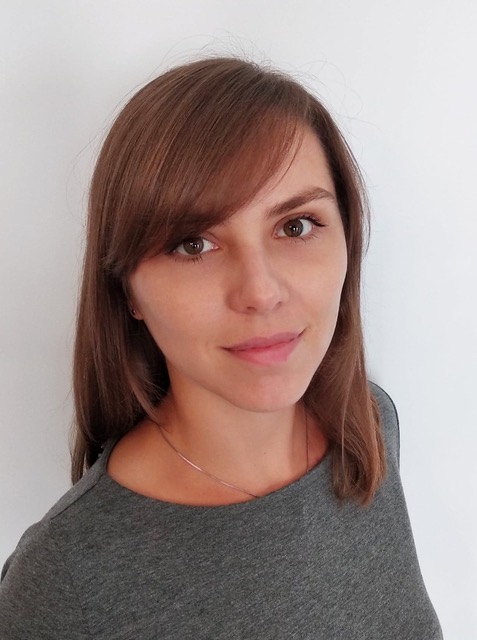}
}
]{Koryna Lewandowska}
is an Assistant in the Faculty of Management and Social Communication in the Institute of Applied Psychology, Department of Cognitive Neuroscience and Neuroergonomics at the Jagiellonian University in Kraków. She obtained her Ph.D. in Psychology on the influence of decision bias on visual recognition memory from the Jagiellonian University in 2019. She received her MA in Psychology from the same institution in 2011. Her research experience includes the realization of projects concerning issues from the fields of cognitive psychology, cognitive neuroscience, chronopsychology, and consumer neuroscience. She is a member of the Polish Association for Cognitive and Behavioural Therapy and a lecturer in the College of Economics And Computer Science.

\end{IEEEbiography}

\begin{IEEEbiography}[
{
\includegraphics[width=1in,height=1.25in,clip,keepaspectratio]{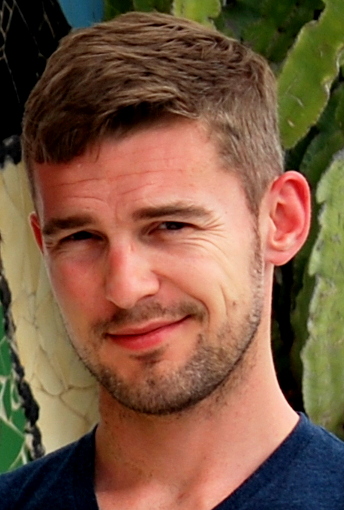}
}
]{Tomasz Trzcinski}
is an Assistant Professor in the Division of Computer Graphics in the Institute of Computer Science at Warsaw University of Technology since 2015. He obtained his D.Sc. degree (habilitation) in 2020 at Warsaw University of Technology and his Ph.D. in Computer Vision at École Polytechnique Fédérale de Lausanne in 2014. He received his M.Sc. degree in Research on Information and Communication Technologies from Universitat Politècnica de Catalunya and M.Sc. degree in Electronics Engineering from Politecnico di Torino in 2010. He is an Associate Editor of IEEE Access and frequently serves as a reviewer in major computer vision conferences (CVPR, ICCV, ECCV, ACCV, BMVC, ICML, MICCAI) and international journals (TPAMI, IJCV, CVIU, TIP, TMM). His professional appointments include work with Google in 2013, Qualcomm Corporate R\&D in 2012, and Telefónica R\&D in 2010. He was a Visiting Scholar at Stanford University in 2017 and at Nanyang Technological University in 2019. He is a co-organizer of warsaw.ai, a member of IEEE and Computer Vision Foundation, an expert of National Science Centre and Foundation for Polish Science, as well as a member of the Scientific Board for PLinML and Data Science Summit conferences. He is a Chief Scientist and Partner at Tooploox, where he leads a team of machine learning researchers and engineers. He co-founded Comixify, a technology startup focused on using machine learning algorithms for editing videos.
\end{IEEEbiography}
\begin{IEEEbiography}[
{
\includegraphics[width=1in,height=1.25in,clip,keepaspectratio]{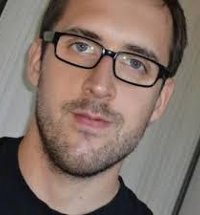}
}
]{Bartosz Zieli{\'n}ski}
is an Assistant Professor in the Faculty of Mathematics and Computer Science in the Institute of Computer Science and Computer Mathematics at Jagiellonian University in Kraków since 2012. He obtained his Ph.D. in Computer Science in the Institute of Fundamental Technological Research at the Polish Academy of Science in 2012. He received his M.Sc. degree in Computer Science from Jagiellonian University in 2007. He frequently serves as a reviewer in international journals on machine learning and medical image analysis (AIR, CSBJ, CBM, T-BME, TRENDS MICROBIOL). His professional appointments include work with Volantis Systems Limited in 2009 and Samsung in 2018. He was a Visiting Scholar at Vienna University of Technology in 2015 and Instituto Superior Técnico in Lisbon in 2019. He is a co-organizer of the Cracow Cognitive Science Conference and Theoretical Foundations of Machine Learning. He is a Lead Data Scientist at Ardigen, where he leads a team of medical image analysis researchers and engineers.
\end{IEEEbiography}

\EOD

\end{document}